\PassOptionsToPackage{table}{xcolor}%
\documentclass{vc3paper}
\usepackage{tikz}
\usetikzlibrary{arrows.meta,positioning,fit}
\usepackage{amsmath}
\usepackage{amssymb}

\usepackage{booktabs}
\usepackage{tabularx}
\usepackage{makecell}
\usepackage[table]{xcolor}

\usepackage{multirow,siunitx,graphicx}
\definecolor{VideoChatGreen}{HTML}{EAF3EA}

\graphicspath{{./figures/}{./assets/}}
\newcommand{\VCArchFigure}[2]{%
  \IfFileExists{figures/#1.pdf}{\includegraphics[width=\linewidth]{figures/#1.pdf}}{%
  \IfFileExists{figures/#1.PNG}{\includegraphics[width=\linewidth]{figures/#1.PNG}}{%
  \IfFileExists{figures/#1.png}{\includegraphics[width=\linewidth]{figures/#1.png}}{%
  \IfFileExists{figures/#1.jpg}{\includegraphics[width=\linewidth]{figures/#1.jpg}}{%
  \IfFileExists{figures/#1.JPG}{\includegraphics[width=\linewidth]{figures/#1.JPG}}{%
  \IfFileExists{assets/#1.pdf}{\includegraphics[width=\linewidth]{assets/#1.pdf}}{%
  \IfFileExists{assets/#1.png}{\includegraphics[width=\linewidth]{assets/#1.png}}{%
  \IfFileExists{assets/#1.PNG}{\includegraphics[width=\linewidth]{assets/#1.PNG}}{%
    \typeout{^^J vc3 WARNING: Missing #1.[pdf/png/jpg] under figures/ or assets/ --- using example-image-#2 ^^J}%
    \includegraphics[width=\linewidth]{example-image-#2}%
  }}}}}}}}%
}%

\AtBeginDocument{%
  \IfFileExists{assets/parrot_logo.png}{%
    \def\VCParrotTitleLogoInline{%
      \nobreak\hspace{0.15em}%
      \raisebox{-0.38ex}{\includegraphics[height=1.2em,keepaspectratio]{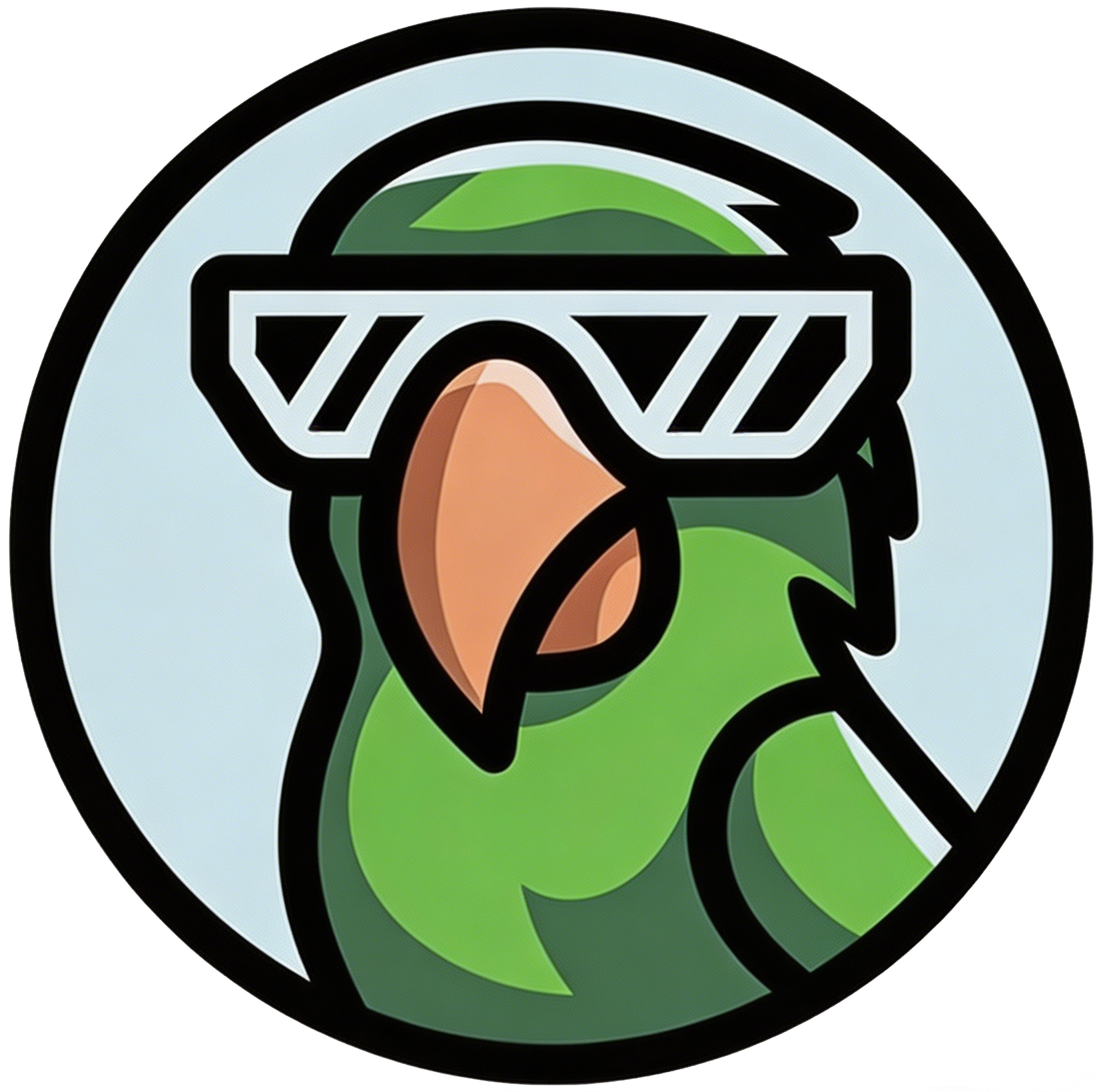}}%
    }%
  }{%
    \def\VCParrotTitleLogoInline{}%
  }%
}

\title{%
  \noindent
  \raggedright
  {\fontsize{24pt}{28pt}\selectfont  \VCParrotTitleLogoInline VideoChat3:}%
  \par\vspace{0.16em}\noindent
  {\fontsize{19.5pt}{23.5pt}\selectfont Fully Open Video MLLM for Efficient and Generalist Video Understanding}

}

\authorOne[1,*]{Xinhao Li}
\authorOne[1,*]{Yuhan Zhu}
\authorOne[1,*]{Xiangyu Zeng}
\authorOne[3,*]{Yuhao Dong}
\authorOne[3,*]{Haoning Wu}
\authorOne[1,*]{Zhiqiu Zhang}
\authorTwo[1]{Yuandong Yang}
\authorTwo[1]{Changlian Ma}
\authorTwo[1]{Qingyu Zhang}
\authorTwo[2]{Yansong Shi}
\authorTwo[1]{Xinyu Chen}
\authorTwo[1]{Haoran Chen}
\authorThree[1]{Zizheng Huang}
\authorThree[1]{Jun Zhang}
\authorThree[4]{Kun Ouyang}
\authorThree[1]{Lin Sui}
\authorThree[2]{Ziang Yan}
\authorThree[2]{Yicheng Xu}
\authorThree[2]{Chenting Wang}
\authorFour[2]{Yinan He}
\authorFour[2]{Hongjie Zhang}
\authorFour[2]{Yi Wang}
\authorFour[2]{Yu Qiao}
\authorFour[2]{Yali Wang}
\authorFour[3]{Ziwei Liu}
\authorFour[2]{Kai Chen}
\authorFour[1,\dagger]{Limin Wang}

\affiliation[1]{Nanjing University}
\affiliation[2]{Shanghai AI Laboratory}
\affiliation[3]{Nanyang Technological University}
\affiliation[4]{Peking University}
\contribution[]{${}^{*}$Equal contribution. ${}^{\dagger}$Corresponding author: \email{lmwang@nju.edu.cn}.}

\abstract{Recent advances in video understanding have spanned motion, long video, and streaming interaction, driving this field toward real-world applications. Despite this progress, current open-source models remain limited in several ways. They often struggle to generalize across diverse video types, making them effective only in specific domains. High computational demands further restrict their efficiency and scalability. Moreover, most models are only partially open, with key components such as training code, strategy, or datasets unavailable, which hinders reproducibility and slows community-driven development. To address these issues, we introduce VideoChat3, a fully open, efficient, and generalist video-centric MLLM. VideoChat3 advances video understanding through two complementary designs. For efficiency, we introduce \textbf{\textit{Inflated 3D Vision Transformer (I3D-ViT)}} and \textbf{\textit{Adaptive Frame Resolution for Streaming Video Perception}}, which enable efficient spatiotemporal representation and reduce the cost of processing video inputs during training and inference. For effectiveness, we develop a scalable video data synthesis pipeline that curates three diverse, high-quality training datasets: \textbf{\textit{VideoChat3-Academic2M}}, \textbf{\textit{VideoChat3-LV116K}}, and \textbf{\textit{VideoChat3-OL617K}}, covering general, long-form, and streaming video scenarios and thereby improving the model's generalization across domains. By integrating these designs, VideoChat3 achieves a rare balance of broad generalization and computational efficiency. Experiments across general, long-form, and streaming benchmarks demonstrate that VideoChat3, despite having only 4B parameters, surpasses prior open-source models with equal or larger parameter counts while achieving higher efficiency. By releasing model weights, training code, training strategy, and complete training datasets, we aim to provide a fully reproducible foundation and help the open-source community bridge gaps in data access and training resources, fostering broader development of real-world video understanding systems. 
}

\metadata[Homepage:]{\url{https://mcg-nju.github.io/VideoChat3}}
\metadata[Model \& Data:]{\url{https://huggingface.co/collections/MCG-NJU/videochat3}}
\metadata[Code:]{\url{https://github.com/MCG-NJU/VideoChat3}}

\begin{document}

\maketitle

\section{Introduction}
\label{sec:intro}

\begin{figure}
    \centering
    \includegraphics[width=1.0\linewidth]{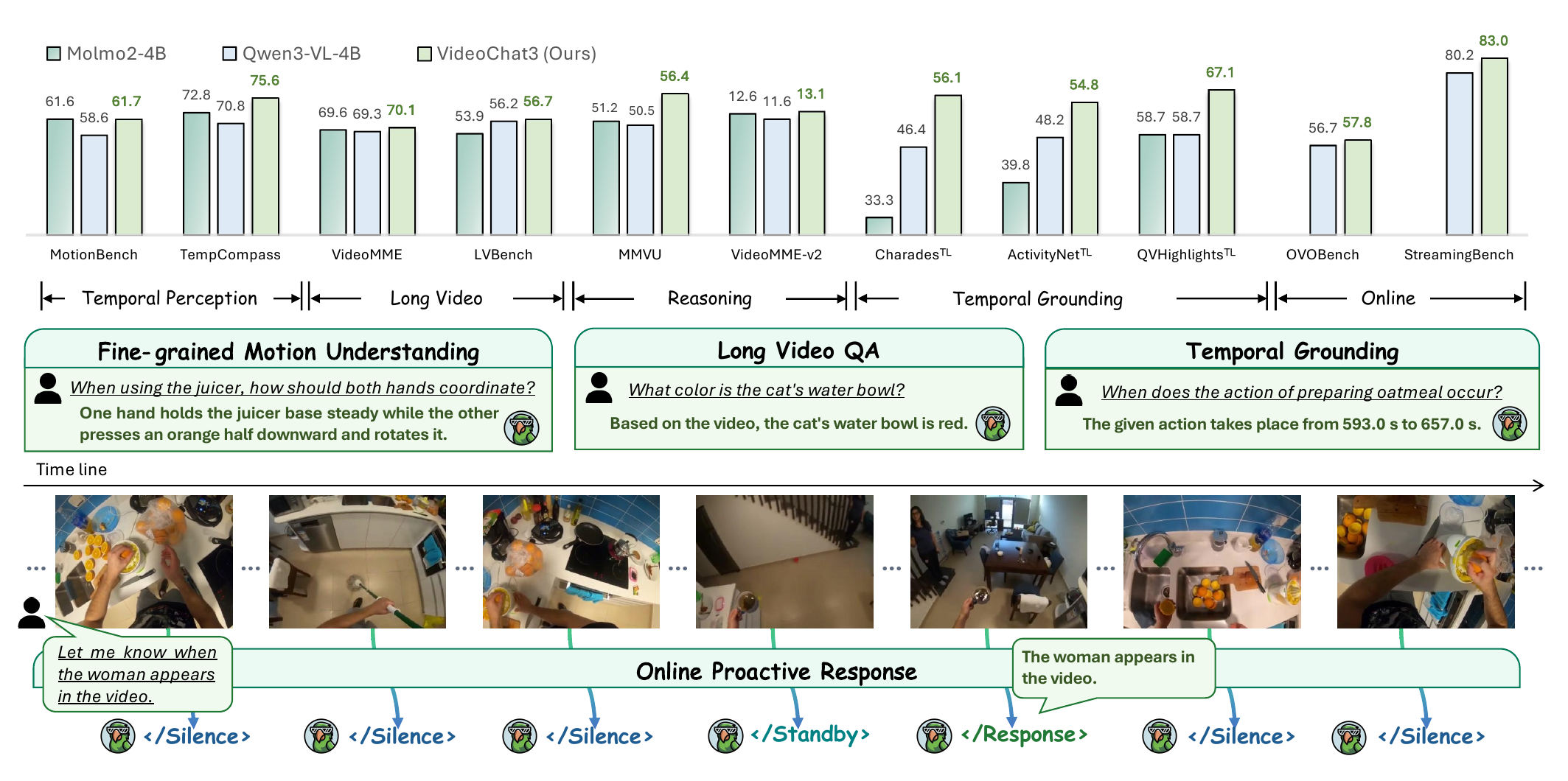}
    \caption{VideoChat3 achieves strong performance across diverse evaluation benchmarks, including temporal perception, long video understanding, and temporal grounding, while also supporting online proactive responses.}
    \label{fig:placeholder}
\end{figure}

Video is the most natural medium through which a human perceives, memorizes, reasons about, and interacts with the physical world. Unlike static images, videos capture continuous visual changes in objects, scenes, actions, and events, requiring models to understand not only spatial semantics but also temporal dynamics, causal evolution, and long-range contextual dependencies. Meanwhile, in the digital era, video has widespread applications, covering short-form social media, long-form entertainment, autonomous driving, embodied perception, surveillance, and real-time human-AI interaction. As a result, building a generalist and efficient video understanding model is a critical step toward general-purpose multimodal intelligence.


Recent progress~\cite{qwen3vl,internvideo2_5,videochat-flash,cho2025PerceptionLM,molmo2} in Video Multimodal Large Language Models (Video MLLMs) has significantly advanced this goal. Built upon large language models~\cite{qwen2,qwen3}, Video MLLMs have demonstrated promising capabilities across fine-grained video perception~\cite{cho2025PerceptionLM,videochat-r1_5}, long-video understanding~\cite{videochat-flash,videoxl2,video-o3}, egocentric perception~\cite{spacer}, temporal grounding~\cite{timechat,timesuite,timelens}, and streaming video understanding~\cite{videollmonline,flashvstream,ovbench-videochatonline,dispider,odvbench-streamforest,streamo,streambridge}. At the same time, commercial multimodal systems such as Gemini~\cite{gemini3}, Kimi~\cite{kimi-k2_5}, and Seed~\cite{seed1_8} have shown increasingly strong video understanding performance, revealing the potential of large-scale video-language modeling. 


However, despite these advances, we identify three key limitations in existing Video MLLMs. First, many models remain specialized for particular video settings or benchmark categories. A model optimized for short videos may not transfer well to long-form understanding, while methods designed for offline video understanding often lack the ability to handle streaming interaction. This limited generalization prevents video MLLMs from being reliably deployed in real-world scenarios, where video inputs vary widely in duration, frame rate, viewpoint, temporal structure, and interaction mode. Second, video understanding is inherently computationally demanding. Naively extending image-based MLLMs to videos leads to a rapid increase in visual tokens, making long-video and real-time streaming understanding prohibitively expensive. Without efficient visual encoding, video MLLMs are difficult to scale, train, and deploy. Third, reproducibility remains a major bottleneck. Several high-performing models are either closed-source or only partially open, with undisclosed training data, incomplete training recipes, or unavailable data construction pipelines. This opacity makes it difficult for the community to understand what drives performance, compare methods fairly, or build upon prior work. All these limitations indicate that the field still lacks an efficient, generalist, and fully reproducible video MLLM that can serve as a strong foundation for open research.

To address these challenges, we introduce VideoChat3, a systematic framework that advances video understanding along three dimensions: efficient model architecture, scalable data construction, and full-stack open-sourcing for reproducible research.

\begin{itemize}
    \item \textbf{Model Architecture}: Starting from the core problem of video encoding, we propose the \textit{Inflated 3D Vision Transformer (I3D-ViT)} architecture together with an \textit{Adaptive Frame Resolution} mechanism for streaming video perception. This design significantly reduces the number of visual tokens in high-frame-rate, long-form, and real-time streaming video scenarios, thereby substantially improving video processing efficiency while maintaining strong understanding accuracy. It effectively alleviates the computational bottleneck faced by existing Video MLLMs in long-video and real-time settings.
    \item \textit{\textbf{Data Construction}}: We develop a large-scale multimodal data synthesis and curation pipeline spanning both offline and online streaming video understanding tasks, yielding three dedicated datasets: \textit{VideoChat3-Academic2M}, \textit{VideoChat3-LV116K}, and \textit{VideoChat3-OL617K}, which collectively comprise 3 million high-quality multimodal instruction samples. Meanwhile, we adopt a multi-stage curriculum learning strategy for progressive model training, enabling the model to systematically acquire video understanding capabilities across different task types and difficulty levels, ultimately leading to strong generalist video understanding performance.
    \item \textbf{Training, Evaluation, and Fully Open-source}: Building upon the synergy between the proposed architecture and data pipeline, VideoChat3 achieves dual breakthroughs in both effectiveness and efficiency. Extensive evaluations on mainstream video understanding benchmarks demonstrate that VideoChat3 surpasses open-source state-of-the-art models of comparable parameter scale, such as Qwen3-VL~\cite{qwen3vl} and Molmo2~\cite{molmo2}, while also showing significant advantages in training throughput and inference speed. More importantly, we fully release the model weights, training code, training strategy, complete training datasets, and dataset construction pipeline, providing a high-performance, efficient, and reproducible open-source foundation for future research in real-world video understanding.
\end{itemize}


Overall, VideoChat3 serves not only as a strong video MLLM, but also as a fully reproducible foundation that lowers the barriers to data, training, and efficient video modeling for the open-source community.

\section{Model Architecture}
\label{sec:architecture}

\subsection{Overview}


Video MLLMs face a fundamental tension between visual perception quality and computational efficiency. As video duration, frame rate, and spatial resolution increase, the number of visual tokens quickly exceeds the context and compute budgets of practical LLM deployments. Most existing systems inherit an image tokenizer and adopt sparse sampling to encode videos into tokens: each sampled frame is encoded largely as an independent image, while temporal modeling is deferred until the resulting visual tokens reach the LLM. This paradigm suffers from two prominent drawbacks. First, sparse frame sampling leads to the direct loss of abundant video details, such as short-term motion and subtle temporal variations. Second, temporally adjacent frames contain considerable overlapping and redundant content, yet the image tokenizer fails to leverage such repetitions to derive more compact video representations. 

To this end, VideoChat3 is built on a different principle: \textit{\textbf{the spatiotemporal structure of videos should be modeled and their redundant information reduced as early as possible before the video tokens are passed to the LLM}}. We introduce two complementary designs that compress video representations along the temporal and spatial dimensions.

\begin{enumerate}
    \item For the temporal dimension, we introduce \textbf{Inflated 3D Vision Transformer (I3D-ViT)}, a visual tokenizer inflated from a pretrained image tokenizer by extending its 2D spatial self-attention into 3D spatiotemporal self-attention. Rather than treating frames as isolated images, I3D-ViT groups consecutive frames into short chunks, applies spatiotemporal self-attention within each chunk for local spatiotemporal modeling, and finally performs temporal pooling to produce compact, motion-aware visual tokens.

    \item For the spatial dimension, we further introduce \textbf{Adaptive Frame Resolution} for streaming video perception. Humans allocate visual attention adaptively: routine moments are perceived coarsely, whereas important or uncertain events prompt finer-grained perception. Inspired by this behavior, our method enables the model to adjust the spatial resolution of incoming frames based on their perceived importance. Thus, uninformative intervals can be processed efficiently at low resolution, while potentially critical moments are examined with richer visual detail.

\end{enumerate}

Together, these designs form a unified video-oriented perception architecture for VideoChat3, enabling efficient and detail-preserving video understanding across diverse deployment scenarios.

\subsection{Efficient Video Tokenization with Spatiotemporal Modeling}

\begin{figure}[t]
  \centering
 \includegraphics[width=1\linewidth]{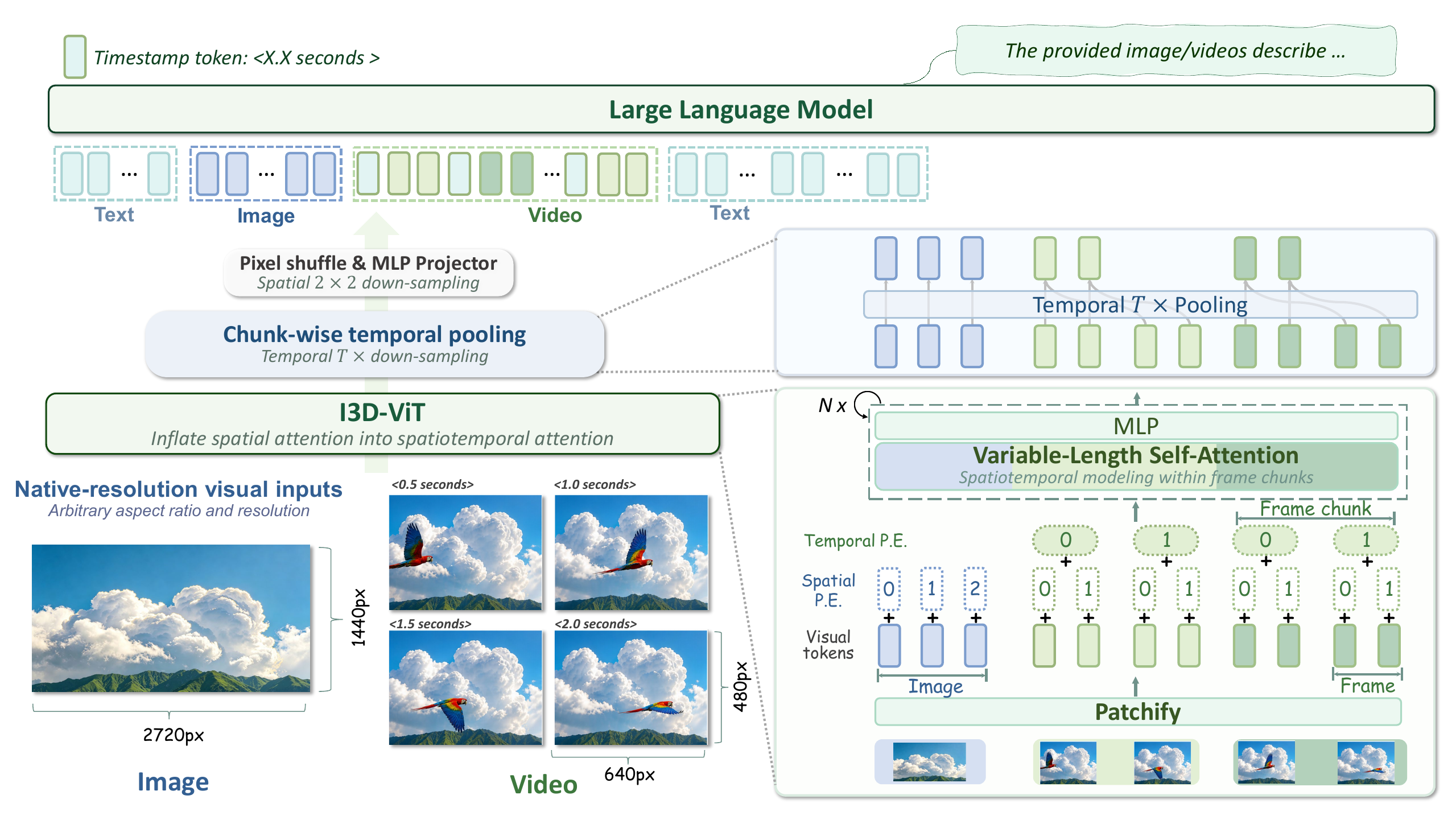}
  \caption{\textbf{VideoChat3 architecture with I3D-ViT.} VideoChat3 follows the classical ViT--MLP Projector--LLM architecture, with I3D-ViT enabling efficient video encoding before visual tokens are passed to the LLM. Specifically, spatial $2\times2$ merging and temporal $T$-frame pooling reduce the visual sequence length by approximately a factor of $4T$. Under the default setting of $T=4$, this yields a $16\times$ spatiotemporal compression ratio; for clarity, the figure illustrates the mechanism with $T=2$.}
  \label{fig:method2-i3d-vit}
\end{figure}


Video is inherently information-rich: both fine-grained short-term motion and long-range video understanding require a large number of visual tokens. However, practical LLMs can only accommodate a limited visual token budget. Existing systems typically address this tension by discarding information before it ever reaches the model, for example by sparsely sampling frames and encoding each sampled frame at a modest resolution. While this keeps the context length manageable, it fundamentally removes substantial video details at the input stage. Our key observation is that local temporal dynamics and inter-frame redundancy can be modeled more efficiently within the visual tokenizer. As illustrated in \Cref{fig:method2-i3d-vit}, we start from a pretrained image tokenizer and inflate its original 2D spatial self-attention into a chunk-wise spatiotemporal attention mechanism. This gives rise to the visual tokenizer of VideoChat3, termed \textbf{Inflated 3D Vision Transformer (I3D-ViT)}. Specifically, I3D-ViT introduces the following designs:

\begin{itemize}
\item \textbf{Chunked Frame Grouping.}
Instead of encoding sampled frames independently, we partition each video into contiguous frame chunks of up to $T$ frames. Each frame chunk is treated as a local spatiotemporal unit, allowing neighboring frames to interact before the resulting visual tokens are passed to the LLM.

\item \textbf{Temporal Positional Encoding.}
Within each frame chunk, we preserve the pretrained spatial positional embeddings from the image tokenizer and introduce learned absolute temporal embeddings for frame indices $0$ through $T-1$, allowing the tokenizer to distinguish the temporal order of frames while retaining the pretrained spatial structure.

\item \textbf{Native-Resolution Spatiotemporal Modeling.}
Tokens from all frames within a frame chunk are flattened into a single sequence, enabling the I3D-ViT blocks to perform joint attention over space and time at the native input resolution. This allows I3D-ViT to produce more compact video tokens after local temporal redundancy has been implicitly modeled by spatiotemporal self-attention.

\item \textbf{Chunk-Wise Temporal Pooling.} 
After spatiotemporal contextualization, we aggregate features along the temporal dimension within each frame chunk. This reduces the token count by a factor of $T$, allowing the subsequent module to operate on features that already encode local temporal redundancy, rather than on frame-wise independent representations.
\end{itemize}

Unless otherwise specified, we set $T=4$, which, combined with the $2\times2$ spatial downsampling induced by pixel shuffling~\cite{internvl1.5}, yields \textbf{an overall 16$\times$ spatiotemporal compression ratio}.
Beyond improving context efficiency, I3D-ViT also provides a flexible interface for variable-length videos. Since tokenization adapts to the available visual budget, inputs can be rescaled according to their estimated load: shorter clips can be represented with higher spatial and temporal fidelity, while long-running streams or hours-long videos can be processed at lower resolution and frame rate. This makes the visual bottleneck adaptive rather than fixed, enabling denser video evidence to be preserved under the same LLM context budget.

\subsection{Adaptive Frame Resolution for Streaming Video Perception}

\begin{figure}[t]
  \centering
   \includegraphics[width=1\linewidth]{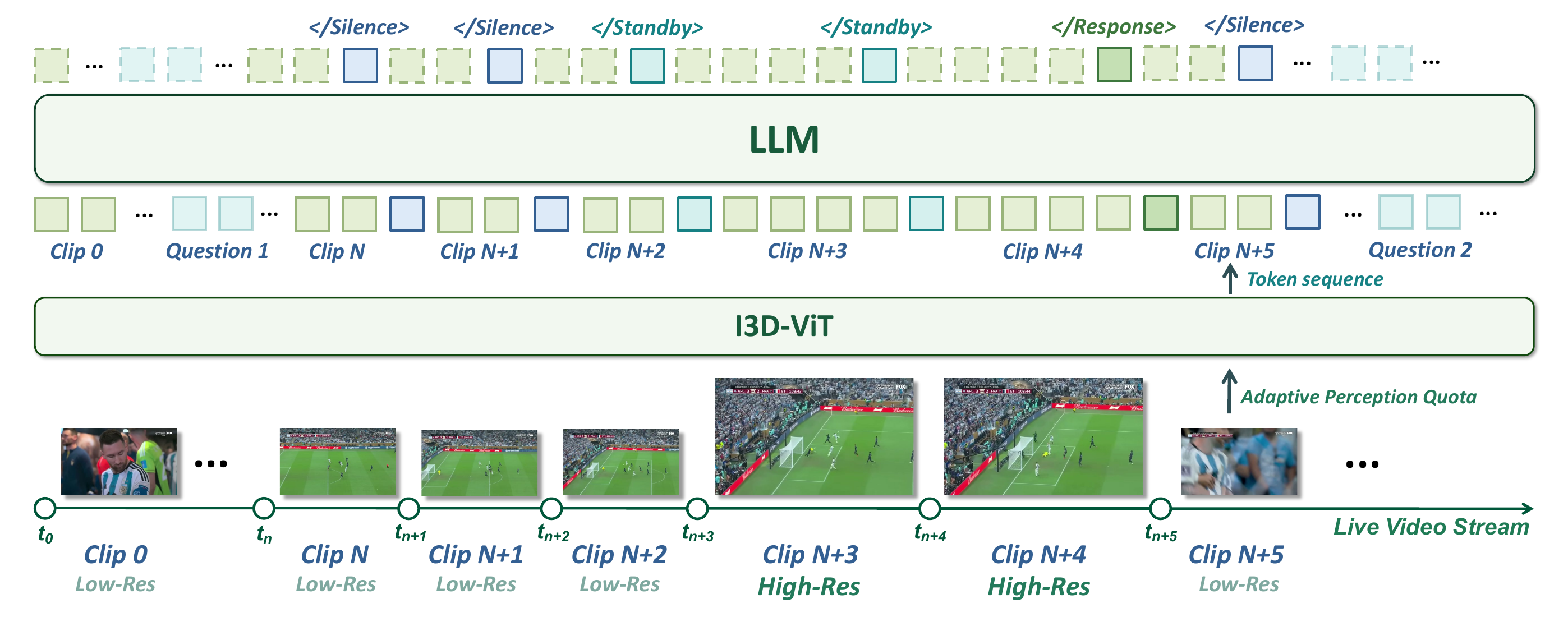}
  \caption{\textbf{Illustration of the Adaptive Frame Resolution Mechanism}. State tokens control whether the next streaming window is encoded at a low or high pixel quota. In the soccer example, routine play is monitored at low cost; when players rush toward the goal, the Standby state enlarges the next window so the model can inspect whether the ball goes in before triggering Response.}
  \label{fig:method1-dynamic-stream}
\end{figure}


Offline video understanding usually assumes that all frames are observed before the model starts decoding. This assumption is unsuitable for streaming scenarios, where visual inputs arrive continuously and the model must decide not only \emph{what} to answer, but also \emph{when} to answer and \emph{how much} visual detail to consume before answering. In this process, the model may need to process visual information at every moment, leading to substantial computational overhead. Therefore, a more efficient streaming video perception scheme is needed.

Our design is motivated by how humans watch live video streams. Viewers rarely inspect every moment with the same level of attention: routine intervals are monitored in a relaxed mode, while potential highlights immediately trigger focused observation. For example, during a soccer broadcast, midfield passing can be followed with coarse attention, but once an attacker breaks through the defense and approaches the goal, the viewer naturally concentrates on the fine visual details needed to judge whether the ball goes in. This suggests that a streaming video model should also maintain low-cost monitoring by default, allocate higher visual fidelity only when a possible response-worthy cue appears, and generate an answer after sufficient evidence has been observed. We therefore formulate streaming inference as a state-conditioned closed loop between the language model and the visual tokenizer, as illustrated in \Cref{fig:method1-dynamic-stream}. At each streaming step, the model first emits a response-state token. The token belongs to one of three states:
\begin{itemize}
  \item \textbf{Silence.} The current window does not contain task-relevant evidence. The model suppresses natural-language generation and continues monitoring the stream.
  \item \textbf{Standby.} The current window contains potential evidence, but the information is not yet sufficient for a reliable answer. The model keeps silent while preparing to observe subsequent windows at higher fidelity.
  \item \textbf{Response.} The accumulated evidence is sufficient. The model switches from state prediction to autoregressive answer generation, and then returns to low-cost monitoring unless a new cue is detected.
\end{itemize}

The key distinction from prior response-state formulations is that the state token is also used as a control signal for the next visual input. Let \(s_t\) denote the state predicted after processing the current video window, and let \(b_{t+1}\) denote the per-frame pixel quota used for the next window. We use a simple deterministic controller:
\[
b_{t+1} =
\begin{cases}
B_{\mathrm{low}}  & \text{if } s_t = \mathrm{Silence},\\
B_{\mathrm{high}} & \text{if } s_t = \mathrm{Standby},\\
B_{\mathrm{low}}  & \text{if } s_t = \mathrm{Response},
\end{cases}
\qquad
B_{\mathrm{low}} = 224^2\ \text{pixels},\quad B_{\mathrm{high}} = 448^2\ \text{pixels}.
\]
Thus, most background frames and post-response monitoring frames are processed with a compact low-quota perception window under a \(224^2\)-pixel per-frame quota, while windows following Standby cues are enlarged under a \(448^2\)-pixel quota to recover fine-grained details such as small objects, subtle motion, or text. Each frame is isotropically resized according to its original aspect ratio, producing a flexible resolution whose total pixels fit within the selected quota, e.g., a \(5{:}3\) frame can be encoded as \(280{\times}168\) under the low quota. Since I3D-ViT naturally supports variable spatial input sizes, this policy does not require a separate high-resolution encoder; it only changes the pixel quota of the next streaming chunk while allowing flexible input resolutions within the corresponding quota. The result is an adaptive dynamic perception window that allocates visual tokens along the time axis according to the model's own uncertainty and evidence state.

\section{Dataset Construction}
\label{sec:data}

\subsection{Overview}
The central challenge in building video instruction data is that no single data source provides both trustworthy supervision and sufficiently rich temporal structure. Existing academic datasets offer broad task coverage and clear source provenance that are often human verified. However, many of these datasets were originally designed for short-clip or closed-form prediction, where the desired output is a class label, an option letter, or a short phrase. This creates a supervision mismatch: the model observes a long multimodal context, but receives only a small number of target tokens, with little explicit evidence guidance.

We therefore treat academic datasets as reliable semantic anchors rather than definitive supervision targets. Starting from their original annotations, we rewrite and validate the labels to yield \textbf{denser supervisory signals paired with evidence-grounded responses}. Our goal is not to alter the task definition or inject unconstrained external knowledge, but to make explicit the visual evidence and reasoning logic implicit in the original labels. This annotation enhancement preserves the reliability of curated academic resources while improving their efficacy for training models in natural-language response generation, visual grounding, and evidence-aware reasoning.

The second challenge is \textbf{long video understanding}. Academic resources are often dominated by short videos and benchmark-style tasks, whereas long-video understanding requires a different kind of supervision. Relevant evidence may appear sparsely across minutes or hours, events may only become meaningful after earlier context, and correct answers often require linking entities, actions, and scene transitions across multiple temporal segments. To cover this regime, we further collect long videos from external sources and build a dedicated synthesis pipeline. The pipeline decomposes each video into coherent temporal units, annotates and filters each unit, and then composes the validated evidence into training annotations. 

Finally, we convert high-quality video QA pairs into online streaming samples, so the model learns not only how to answer properly, but also how to identify the correct response timing and proactively respond to user needs once sufficient visual evidence has emerged.

\subsection{VideoChat3-Academic2M: Re-Annotating Academic Video Data}

\begin{figure}[t]
  \centering
 \includegraphics[width=1\linewidth]{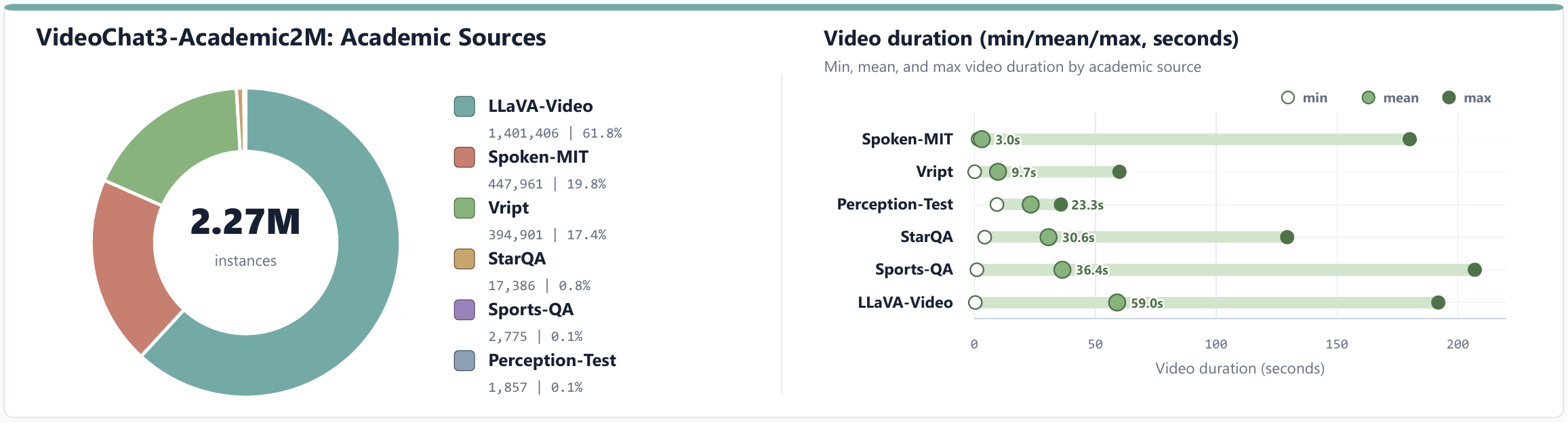}
\caption{\textbf{Source Distribution of VideoChat3-Academic2M.} VideoChat3-Academic2M contains 2.27M caption/QA instances from six academic sources, dominated by LLaVA-Video~\cite{llava_video}, Spoken-MIT~\cite{S-MiT}, and Vript~\cite{Vript}.}
  \label{fig:academicdata}
\end{figure}

\paragraph{Academic Source Aggregation.}
As shown in \Cref{fig:academicdata}, we first aggregate publicly available academic datasets covering video captioning, video question answering, and fine-grained motion perception. These datasets provide complementary supervision signals: captioning data describes visible entities and actions, QA data encourages query-conditioned perception, and motion-centric data stresses fine-grained temporal understanding. More importantly, they are curated under explicit task definitions and often contain human-verified labels, making them reliable sources of visual semantics.
Nevertheless, their original annotation formats are not always suitable for instruction tuning. A large portion of samples only provide a multiple-choice option, a single noun phrase, or a short factual answer. Directly training on such annotations gives the model little generative supervision relative to the number of video and prompt tokens. The model may learn the final label, but it receives limited signal about how the answer should be expressed, what visual evidence should be mentioned, or how the response should be grounded in the video.

\begin{figure}[t]
  \centering
 \includegraphics[width=1\linewidth]{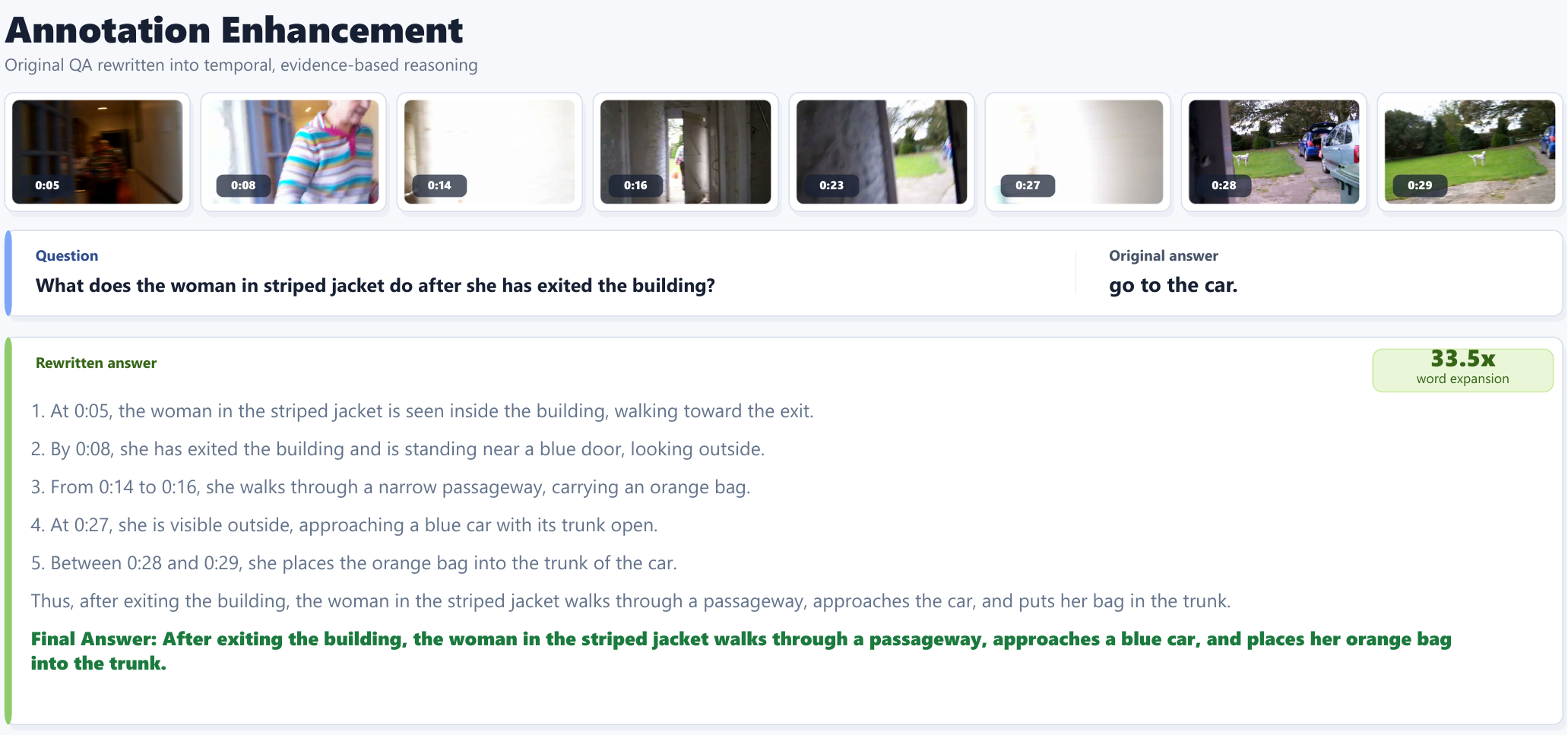}
  \caption{\textbf{Example of Annotation Enhancement from Concise Answers to Evidence-Grounded Responses.}
A short-phrase answer is rewritten into a temporally grounded, evidence-rich response that explains the observed actions and supports the final answer with video-specific cues. This illustrates how annotation enhancement increases supervision density while preserving the original semantic label.}
  \label{fig:academicdataexample}
\end{figure}

\paragraph{Evidence-Grounded Annotation Enhancement.}
To better exploit these reliable but sparse academic annotations, we enhance them into richer instruction-following responses. The key principle is to densify the supervision while keeping the original answer as a semantic constraint. For each sample, we first normalize the query by removing evaluation-specific formatting requirements, such as prompts that force the model to output only an option letter. When the original response is option-based, we recover the natural-language content of the selected option and use it as the target semantics.
We then use Qwen3-VL-235B-A22B to rewrite short-text or option-only responses into comprehensive answers with explicit visual evidence and reasoning. The model is instructed to preserve the original answer, avoid adding unsupported facts, and ground the response in observable video content. In this way, an answer that originally provides only the final decision is transformed into supervision that also describes the relevant objects, actions, scenes, and temporal cues. As shown in \Cref{fig:academicdataexample}, the rewritten annotations contain substantially more video-specific information than the original annotations, providing denser targets for learning video-language alignment and evidence-aware generation.

\paragraph{Consistency Verification and Error Filtering.}
Annotation enhancement also introduces a new risk: a powerful generator may produce fluent but unsupported details. We therefore add a verification stage to preserve the fidelity of the original academic labels. Specifically, we use Qwen3-VL-235B-A22B as a judge model to compare each rewritten response with the original annotation. The judge checks whether the rewritten response is semantically consistent with the original label and whether the added explanation changes, contradicts, or over-specifies the answer. Only annotations that pass this consistency check are retained.
This filtering step lets us increase the density and readability of supervision without sacrificing the reliability that makes academic datasets valuable in the first place. In effect, the original annotation defines the semantic boundary of the answer, while the rewritten response supplies a richer and more trainable expression of that answer.

\subsection{VideoChat3-LV116K: Long-Video Data Synthesis}

\begin{figure}[t]
  \centering
 \includegraphics[width=1\linewidth]{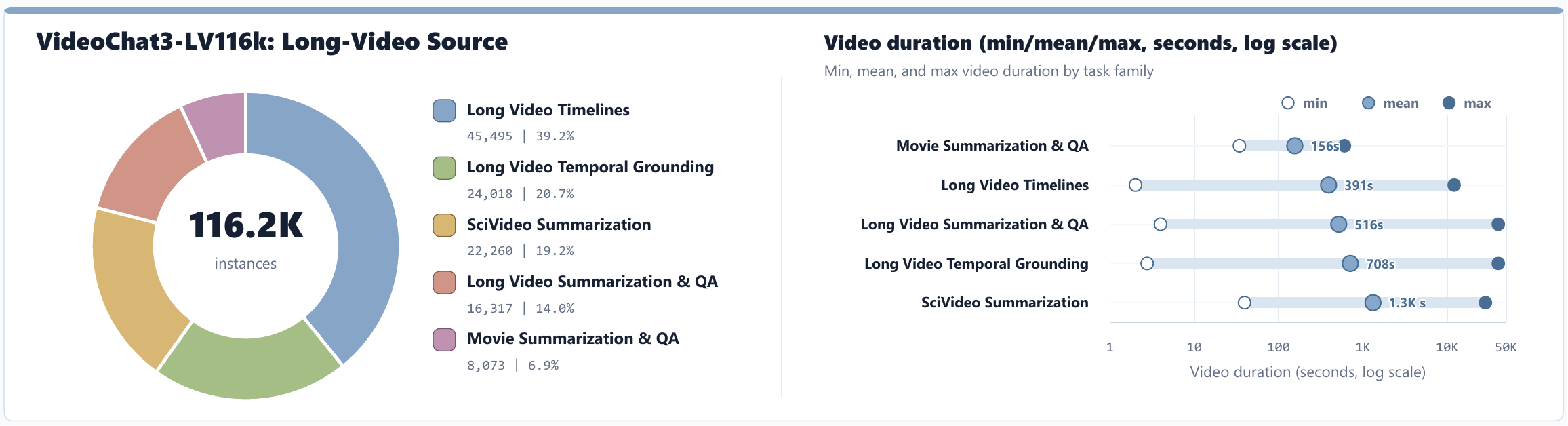}
  \caption{\textbf{Source Distribution of VideoChat3-LV116K.} The panels summarize the collected long-video repository used to construct VideoChat3-LV116K, with 116.2K rows. The duration statistics reveal a clear temporal-scale gap: academic sources are mostly short clips, with mean durations from 3.0s to 59.0s, whereas our collected long-video shards average 156s to 1.3K seconds and extend to much longer maxima. This complementarity provides both reliable short-clip semantic anchors and long-range supervision for sparse evidence, cross-segment aggregation, and event-level reasoning.}
  \label{fig:lvcdata}
\end{figure}

\begin{figure}[t]
  \centering
 \includegraphics[width=1\linewidth]{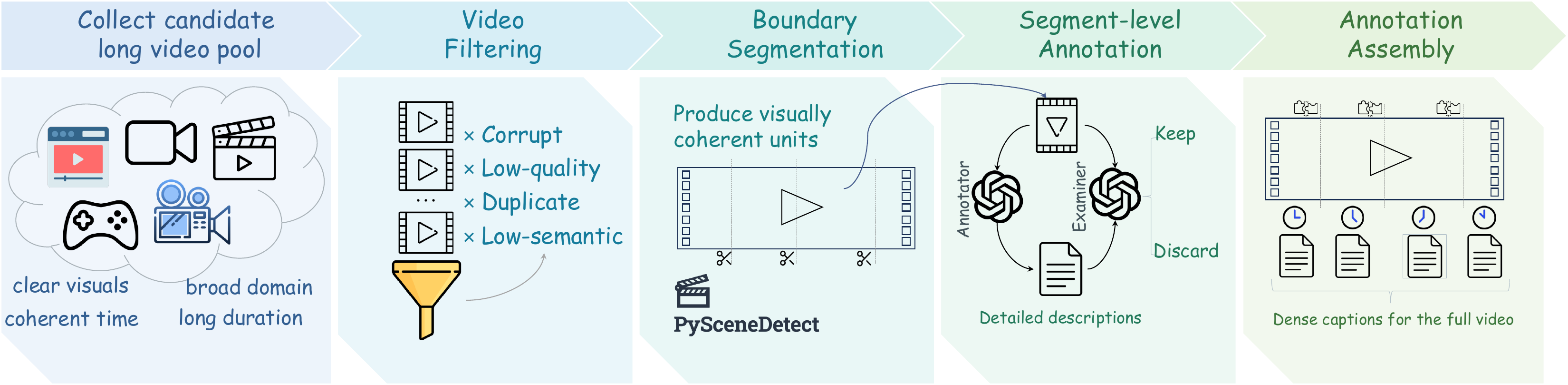}
  \caption{\textbf{Long-Video Data Synthesis Pipeline for VideoChat3-LV116K.}
The pipeline converts raw long videos into structured supervision through candidate filtering, temporal boundary segmentation, segment-level annotation, quality examination, and full-video annotation assembly. Instead of annotating an entire long video directly, it builds a validated segment-level evidence ledger, making sparse long-range events easier to capture and reason over.}
  \label{fig:data-construction-pipeline}
\end{figure}

Although enhanced academic data improves the quality of short-video and benchmark-style supervision, it does not fully address long-video understanding. Long videos are not simply longer versions of short clips: their difficulty comes from sparse evidence, delayed dependencies, event transitions, and information that must be aggregated over extended temporal contexts. A model trained only on short clips may recognize local actions, but still fail to connect events that are separated in time or to localize evidence precisely within a long sequence. To address this gap, we construct VideoChat3-LV116K by collecting additional long videos and synthesizing structured annotations through a multi-stage pipeline. As illustrated in~\Cref{fig:data-construction-pipeline}, the pipeline filters candidate long videos, segments them into coherent temporal units, annotates each segment with quality control, and assembles the validated segment descriptions into dense full-video captions.

\paragraph{Long-Video Collection and Filtering.}
We collect candidate long videos from multiple sources, prioritizing those with clear visual content, coherent temporal structures, and sufficient durations for long-context reasoning. The dataset encompasses a broad range of domains, including entertainment, education, sports, news, science, and gaming. Videos exhibiting severe corruption, low visual quality, excessive duplication, or limited semantic content are discarded. This newly collected data is intended to complement existing academic datasets by supplying extended temporal scales and richer event structures, while established academic resources maintain reliable task diversity.

\paragraph{Boundary-Aware Temporal Segmentation.}
Directly annotating an entire long video is costly, noisy, and constrained by the context limits of current video-language models. We therefore decompose each video into manageable temporal segments before annotation. This segmentation serves as an intermediate representation: each segment is short enough to be described accurately, while the ordered list of segments preserves the global temporal structure needed for long-video reasoning. Segment boundaries are determined by combining visual scene detection with heuristic temporal windows. Specifically, we use PySceneDetect to identify shot transitions and enforce minimum and maximum duration constraints to avoid both overly fragmented clips and segments that are too long for reliable annotation. This produces visually coherent units with controllable temporal granularity.

\paragraph{Segment-Level Annotation and Quality Control.}
Each segment is annotated independently before being integrated into long-video supervision. Using Qwen3-VL-235B-A22B~\cite{qwen3vl}, we generate structured descriptions that capture visible entities, actions, scenes, event transitions, salient temporal details, and OCR or subtitle cues when available. These timestamped descriptions form an evidence ledger for the full video: instead of asking a model to reason over raw long video at once, we first build a validated textual representation of what happens and when it happens.
To reduce hallucination and annotation noise, an auxiliary MLLM evaluates the generated segment descriptions. We discard annotations that are repetitive, overly generic, internally inconsistent, or unsupported by the visual evidence. The remaining descriptions are then aggregated with their corresponding temporal boundaries to form dense, timestamped captions for the full video.

\paragraph{Long-Video Training Annotation Synthesis.}
After obtaining validated segment-level descriptions, we synthesize long-video training labels that require reasoning beyond isolated clips. We query a frontier LLM with an interleaved sequence of timestamps and segment descriptions, prompting it to reason over the entire video context and explicitly cite supporting temporal evidence. This turns the segment evidence ledger into diverse instruction data for the following task families:

\begin{enumerate}
    \item \textbf{Long Video Temporal Grounding.} The model must localize the time intervals corresponding to a given text query. Unlike existing temporal grounding data that only focuses on single-interval localization, we also synthesize queries for multi-interval grounding. We prompt the LLM to generate queries at multiple levels of specificity, including keywords, phrases, and complete sentences, together with candidate temporal regions that may contain the relevant evidence. An expert grounding model is then applied within each candidate region to produce precise temporal spans. We merge spans across the full video, discard empty predictions, and filter samples with low semantic similarity between the query and the grounded video content.
    \item \textbf{Long Video Timelines.} The model is required to describe salient events together with their temporal boundaries. We use the validated segment descriptions to identify event-level units, merge adjacent segments when they describe the same continuous event, and generate concise captions that summarize what happens in each time interval. Samples are filtered when the caption is too generic, temporally ambiguous, or insufficiently supported by the segment evidence.
    \item \textbf{Long-Video QA.} The QA tasks are designed so that the answer depends on evidence distributed across multiple temporal segments. We prompt the LLM to generate questions, answers, and supporting timestamps from the full sequence of segment descriptions. To avoid trivial examples, we filter out questions that can be answered from a single frame, a single isolated segment, or language priors alone. The retained samples encourage the model to aggregate information over long contexts and to provide answers grounded in specific video moments.
    
\end{enumerate}

Finally, we apply an additional filtering mechanism to remove samples that can be solved without meaningful video understanding, such as questions answerable from a single static frame or from common language priors. This ensures that the synthesized annotations emphasize the capabilities that long videos uniquely require: temporal localization, cross-segment aggregation, and event-level reasoning. 

The resulting VideoChat3-LV116K data contains 116.2K JSONL rows from six long-video shards, as shown in \Cref{fig:lvcdata}. While the academic sources have mean durations ranging from 3.0s to 59.0s, the final long-video shards have mean durations from 156s to about 1.3K seconds. VideoChat3-LV116K complements VideoChat3-Academic2M not merely by adding more samples, but by providing supervision over much longer temporal contexts, where evidence is sparse, events are distributed across segments, and answers require cross-event aggregation.

\subsection{VideoChat3-OL617K}

\begin{figure}[t]
  \centering
 \includegraphics[width=1\linewidth]{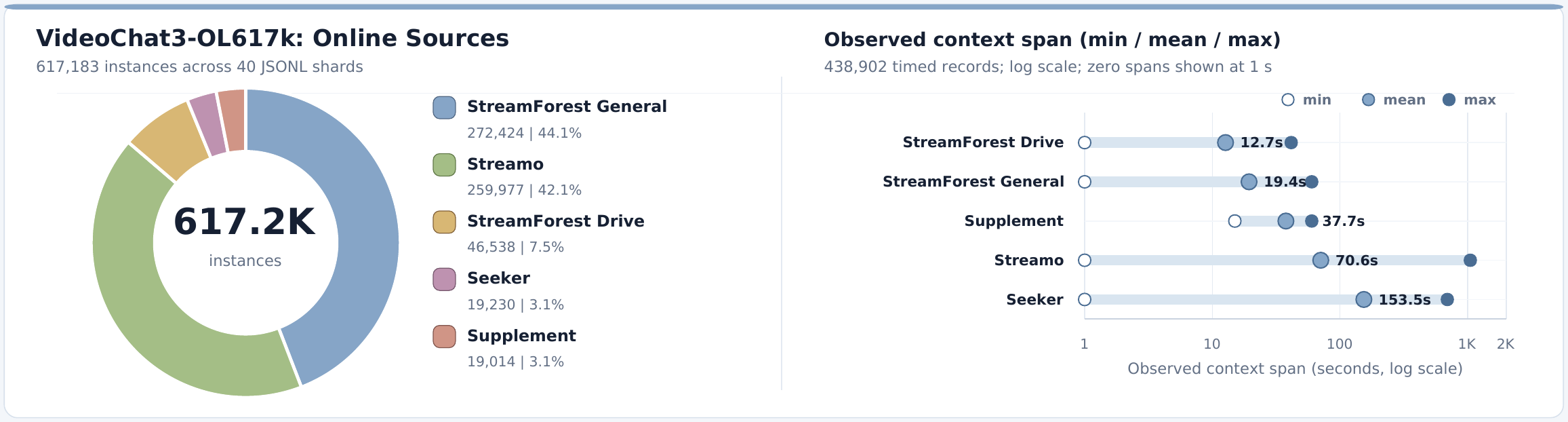}
  \caption{\textbf{Source composition and temporal coverage of VideoChat3-OL617K.} \textit{Left:} Distribution of 617,183 instances across 40 JSONL shards. StreamForest General and Streamo contribute 272,424 (44.1\%) and 259,977 (42.1\%) instances, respectively, while StreamForest Drive, Seeker, and Supplement provide the remaining data. \textit{Right:} Minimum, mean, and maximum observed context spans for 438,902 records with timing information, shown on a logarithmic scale with zero-length spans placed at 1 second for visualization. Mean spans range from 12.7 seconds for StreamForest Drive to 153.5 seconds for Seeker, providing supervision across both short- and long-horizon causal streaming contexts. This diversity supports the evidence accumulation and response-timing behaviors used to construct proactive streaming QA supervision.}
  \label{fig:oldata}
\end{figure}

To train VideoChat3 as a proactive streaming assistant, we convert high-quality video--question--answer triples into online response supervision. Each original sample consists of a video, a user query, and a ground-truth answer. Instead of asking the model to answer after observing the full video, we annotate when the necessary visual evidence becomes available and construct a causal training sequence with explicit response-state tokens.

\paragraph{Visual Clue Localization.}
We first feed the original video--QA triple into a vision language model (VLM) to identify temporal intervals that contain key evidence for answering the query. For each localized interval, the VLM also generates a detailed description of the relevant visual clue, such as the object state, action, text, or event transition that supports the answer.

\paragraph{Clue Verification.}
To reduce noisy or weak localizations, we crop the candidate clue intervals from the original video and ask the VLM to re-evaluate them against the original query. We retain only the samples for which the ground-truth answer can be correctly inferred from the cropped segment alone. This filtering step ensures that the retained intervals provide sufficient visual evidence, rather than merely co-occurring with the answer in the full video context.

\paragraph{Streaming QA Construction.}
After verification, we transform each sample into an interleaved streaming sequence of video windows and target tokens. Verified clue intervals are treated as moments when the model should collect visual evidence, and their corresponding windows are labeled with \texttt{\textless/Standby\textgreater}. Once the clue interval ends, the model is required to immediately emit \texttt{\textless/Response\textgreater} followed by the answer text. All other irrelevant windows are labeled with \texttt{\textless/Silence\textgreater}, encouraging the model to suppress premature responses while continuously monitoring the stream. This process turns offline video QA supervision into proactive online QA data with explicit evidence acquisition and response timing signals.

\section{Training}
\label{sec:train}

The overall training recipe proceeds through four stages, gradually moving from visual tokenizer pre-training to video-language alignment, general video instruction tuning, and finally long-form streaming adaptation. This staged design allows the model to first acquire robust visual representations, then align them with the language space, and eventually specialize in instruction-following scenarios with increasingly complex temporal requirements. \Cref{tab:train-stage-config} summarizes the scheduling and optimization settings for all stages.

\begin{table}[t]
    \centering
    \caption{
        Detailed configuration for each training stage of VideoChat3.
        Dataset items denote the total number of samples used across all
        sub-stages within each stage.
        The corresponding training curricula are described in the subsections below.
    }
    \label{tab:train-stage-config}
    \footnotesize
    \setlength{\tabcolsep}{5pt}
    \renewcommand{\arraystretch}{1.25}
    \definecolor{HeaderBlue}{HTML}{EAF2FF}
    \definecolor{RowGray}{HTML}{F7F8FA}

    \begin{tabularx}{\linewidth}{
        @{}
        >{\raggedright\arraybackslash\bfseries}p{0.20\linewidth}
        *{4}{>{\centering\arraybackslash}X}
        @{}
    }
        \toprule
        \textbf{Stages}
        & \textbf{Stage 0}
        & \textbf{Stage 1}
        & \textbf{Stage 2}
        & \textbf{Stage 3} \\
        \midrule

        Purpose
        & \makecell[c]{Visual tokenizer\\pre-training}
        & \makecell[c]{Video--language\\alignment}
        & \makecell[c]{Video instruction\\tuning}
        & \makecell[c]{Long \& streaming\\instruction tuning} \\

        \rowcolor{RowGray}
        Batch size
        & 512 & 256 & 256 & 256 \\

        Learning rate
        & $1{\times}10^{-3}{\rightarrow}4{\times}10^{-5}$
        & $1{\times}10^{-3}{\rightarrow}4{\times}10^{-5}$
        & $8{\times}10^{-5}$
        & $2{\times}10^{-5}$\\

        \rowcolor{RowGray}
        Dataset items
        & 7.59M & 3.47M & 10.33M & 3.41M \\

        Packed seq. length
        & 8192 & 16384 & 32768 & 98304 \\

        \rowcolor{RowGray}
        Training epochs
        & 1 & 1 & 1 & 1 \\

        Trainable modules
        & Proj. $\rightarrow$ All
        & Proj. $\rightarrow$ All
        & All
        & Proj. \& LLM \\
        \bottomrule
    \end{tabularx}

    \vspace{2pt}
    \parbox{\linewidth}{\scriptsize
        \textit{Note.}
        The arrow indicates the transition from projector warm-up to joint
        training. The temporary LLM used in Stage~0 is discarded after
        visual-tokenizer pre-training.
    }
\end{table}

\subsection{Stage-0: Video Tokenizer Pre-training}

\paragraph{Training Strategy.}
Before training the full MLLM, we first pre-train the visual tokenizer in a language-grounded manner. The goal of this stage is not to endow the model with high-level instruction-following ability, but to make the visual encoder produce representations that can be readily consumed by an autoregressive language model. To this end, we initialize I3D-ViT from the image-pretrained MoonViT~\cite{kimivl}, attach a lightweight projector, and use Qwen3-4B as a temporary text decoder. The model is trained on image-text and video-text pairs with the standard next-token prediction objective, so that visual tokens are optimized directly under a language modeling supervision signal. This pre-training stage consists of two steps:
\begin{enumerate}
    \item \textbf{Projector Warm-up.}
    We freeze both the ViT and the LLM, and update only the projector. This isolates the initial modality alignment problem from representation drift: the image-pretrained ViT preserves its spatial perception ability, while the projector learns to map visual features into the language embedding space.

    \item \textbf{Full-parameter Fine-tuning.}
    We then unfreeze the ViT, projector, and LLM, and train them jointly on a larger and more diverse mixture. This allows the ViT to adapt from image-centric representations to spatiotemporal video representations, while still being anchored by abundant image-text data.
\end{enumerate}

After Stage-0, we discard the temporary language decoder and retain only the trained ViT weights to initialize the visual component of the subsequent MLLM training stages.

\paragraph{Data Mixture.}
Our data mixture is designed to align with the two-stage training strategy:
\begin{enumerate}
    \item \textbf{Projector Warm-up.}
    We use a compact but balanced set of 1.04M single-turn samples, including 558K image-text pairs from LLaVA-Pretrain~\cite{llava} and 481K video-text caption pairs from S-MiT~\cite{S-MiT}. This mixture provides enough visual diversity for establishing the initial visual-language mapping, while keeping the optimization focused on the projector.

    \item \textbf{Full-parameter Fine-tuning.}
    We scale the mixture to 6.55M samples, consisting of 4.9M image-text samples and 1.65M video-text samples. The image portion remains dominant, serving as a strong spatial-semantic anchor when the ViT parameters are unfrozen. It includes image caption samples from CapRL-2M~\cite{caprl} whose source images come from ShareGPT4V~\cite{ShareGPT4V} and DenseFusion~\cite{DenseFusion-1M}, as well as 2.9M CC3M~\cite{cc3m} samples that we recaptioned with CapRL-3B~\cite{caprl}. The video portion introduces temporal and motion-centric supervision from both curated in-house recaptioned data and public video-text resources. It includes 323K VideoChat3-LV samples from ShareGemini WebVid~\cite{WebVid} and K400~\cite{k400} videos, 448K VideoChat3-Academic samples from Qwen3-VL-rewritten S-MiT~\cite{S-MiT} captions, 395K Vript~\cite{Vript} samples, 104K PE-Video~\cite{Perception-Encoder} samples, and 380K Tarsier-recaptioned~\cite{Tarsier2} public video samples from sources such as WebVid~\cite{WebVid}, LSMDC~\cite{LSMDC}, VATEX~\cite{VATEX}, ActivityNet~\cite{Activitynet}, Charades~\cite{Charades}, Something-Something V2~\cite{sthsthv2}, and Ego4D~\cite{Ego4d}.
\end{enumerate}

Overall, Stage-0 uses 7.59M single-turn image/video-text samples.

\subsection{Stage-1: Video-Language Alignment}

\paragraph{Training Strategy.}
After Stage-0, the I3D-ViT tokenizer has already been trained under language-modeling supervision and can produce language-grounded visual representations. Stage-1 therefore focuses on integrating this tokenizer into the final MLLM, rather than re-learning visual semantics from scratch. In this stage, we replace the temporary decoder used in Stage-0 with the target language model and align the complete video-language pathway. The supervision remains caption-centric: image captions continue to provide dense signals for objects, attributes, OCR, and spatial relations, while video captions strengthen the model's sensitivity to motion, temporal transitions, and event-level semantics.

We follow a lightweight-to-joint alignment schedule:
\begin{enumerate}
    \item \textbf{Projector warm-up.}
    We first freeze both the Stage-0 initialized ViT and the LLM, and update only the projector. Since the visual tokenizer has already acquired language-grounded representations, this step mainly adapts the interface between the tokenizer and the target LLM, preventing the newly initialized projector from destabilizing either side.

    \item \textbf{Joint video-language alignment.}
    We then unfreeze the visual tokenizer together with the LLM and continue training with caption-style supervision. This step allows the visual front-end and the language model to co-adapt under the final MLLM architecture, while still avoiding benchmark-specific instruction formats at this early stage.
\end{enumerate}
After Stage-1, images and videos are represented as compact visual tokens that the LLM can reliably narrate, providing a stable foundation for the later reasoning and instruction-following stages.

\paragraph{Data Mixture.}
The Stage-1 data mixture follows the same caption-oriented principle as Stage-0, but is used for interface alignment within the final MLLM. All samples are single-turn visual prompt--caption response pairs.
\begin{enumerate}
    \item \textbf{Projector warm-up.}
    We use 1.04M caption pairs, consisting of 558K image-caption samples from LLaVA-Pretrain~\cite{llava} and 481K video-caption samples from S-MiT~\cite{S-MiT}. This compact mixture preserves the same image-video balance used for the initial tokenizer warm-up, allowing the projector to learn a stable mapping without shifting the pretrained visual or language representations.

    \item \textbf{Joint video-language alignment.}
    We further train on 2.43M caption pairs. The image portion is dominated by CapRL-2M~\cite{caprl}, including samples whose source images come from ShareGPT4V~\cite{ShareGPT4V} and DenseFusion~\cite{DenseFusion-1M}. The video portion includes 323K ShareGemini~\cite{sharegemini} VideoChat3-LV samples from WebVid~\cite{WebVid} and Kinetics-400~\cite{k400}, together with 104K PE-Video~\cite{Perception-Encoder} samples and 2K ShareGPT-4o~\cite{sharegpt-4o} video-caption samples.
\end{enumerate}
Overall, Stage-1 uses 3.47M single-turn caption samples, including 2.56M image-caption samples and 0.91M video-caption samples.

\subsection{Stage-2: Video Instruction Tuning}

\paragraph{Training Strategy.}
Stage-2 converts the aligned MLLM into a general-purpose video instruction-following model. We unfreeze all model parameters and train with a larger packed sequence length on a balanced mixture of image and video QA/caption samples, corresponding to around 50 billion training tokens. Unlike Stage-1, which mainly teaches the model to describe visual content, this stage exposes the model to question-conditioned evidence selection, open-ended answering, multiple-choice reasoning, and caption generation across both image and video inputs.

The image--video mixture is important for stability. Many video failures originate from weak frame-level perception, such as missing small objects, text, spatial relations, or subtle attributes; retaining high-quality image supervision therefore acts as an anchor for fine-grained spatial understanding. Video supervision then builds on this anchor to strengthen temporal reasoning, event ordering, state-change recognition, and multi-segment evidence aggregation. Captioning and QA also play complementary roles: captioning supplies dense token-level supervision and discourages terse answer-only behavior, while QA forces the model to select task-relevant visual evidence and express it in the form requested by the user. In particular, the rewritten academic annotations provide richer rationales than option-only labels, increasing the amount of useful language supervision per visual context. By postponing full instruction tuning until after caption-based alignment, Stage-2 can spend most of its capacity on higher-level video reasoning and response generation, rather than on repairing raw modality mismatch.

\paragraph{Data Mixture.} Stage-2 training data contains 10.33M samples and 23.62M QAs.
The mixture is intentionally anchored by a large image instruction corpus, Honey-Data~\cite{Honey-Data}, which contributes 5.65M effective image samples and 12.85M effective QA turns. It covers captioning, general VQA, STEM diagrams, charts/tables, document/OCR understanding, and grounding/counting, helping preserve strong still-image perception while the model is adapted to video. We additionally include 1.81M text-only Dolci-Instruct-SFT~\cite{olmo2025olmo3} samples to maintain general instruction following, mathematical reasoning, multilingual behavior, and science-oriented responses.

Video supervision is dominated by VideoChat3-Academic, with 1.54M effective video samples and 5.57M effective QA turns. This source includes academic/youtube-style video captions and QA, Qwen3-VL-rewritten or corrected subsets, and duration-bucketed open-ended and multiple-choice examples; its high QA density provides the main signal for video captioning, temporal comprehension, and answer-style alignment. We also include VideoChat3-LV Gemini-generated long-video data (46.7K samples), which complements the academic mixture with long-form, scientific, and cinematic grounding-style supervision.

The remaining video data are smaller but targeted open-source sources. TimeLens-100K~\cite{timelens}, Molmo2~\cite{molmo2}, TVQA~\cite{TVQA}, FineVideo~\cite{FineVideo}, CinePile-v2~\cite{Cinepile}, and VCG-plus~\cite{VideoGPTplus} provide dense multi-turn temporal and subtitle-aware QA. Vript~\cite{Vript}, PE-Video~\cite{Perception-Encoder}, TGIF~\cite{TGIF-QA}, CLEVRER~\cite{CLEVRER}, Something-Something-v2~\cite{sthsthv2}, EgoIT~\cite{Exo2Ego}, MotionBench~\cite{motionbench}, ActivityNet~\cite{Activitynet}, LaSOT/GOT-10k~\cite{Lasot,GOT-10k}, and related tracking/localization datasets broaden coverage over generic video captioning, action recognition, procedural/egocentric understanding, object tracking, physical/causal reasoning, and fine motion discrimination. This composition balances scale with diversity: image/text data stabilize general multimodal instruction following, while video-specific subsets concentrate supervision on temporal grounding, long-video understanding, and fine-grained motion reasoning.

\subsection{Stage-3: Long \& Streaming Video Instruction Tuning}

\paragraph{Training Strategy.}
In this stage, we further adapt the model to long-form and streaming video understanding, where the model is expected to reason over substantially longer temporal contexts and produce responses under incremental visual inputs. We adopt a smaller learning rate together with an extended context window to stabilize optimization over long sequences, processing approximately 10 billion tokens in total. To avoid biasing the model toward either offline holistic perception or online temporal reasoning, we construct a mixed training curriculum that jointly covers conventional offline videos and streaming-style samples.

For streaming videos, we introduce a dedicated training strategy that exposes the model to progressively available video segments and requires it to answer based only on the information observed so far. This design encourages the model to maintain temporal memory, update its understanding as new frames arrive, and preserve strong performance on standard offline video tasks. As a result, the final stage bridges the gap between long-context video instruction following and practical streaming video interaction.

\paragraph{Supervision Mask for State Transition Learning.}
We train the streaming behavior with teacher-forced state supervision. Each training sample is represented as an interleaved sequence of video windows and target tokens. For non-answer moments, the target token is one of \texttt{\textless/Silence\textgreater} or \texttt{\textless/Standby\textgreater}; when sufficient evidence has appeared, the target token becomes \texttt{\textless/Response\textgreater} and is followed by the answer text. The answer text uses the standard next-token prediction loss. The special treatment is applied only to the response-state tokens.

\begin{figure}[t]
  \centering
   \includegraphics[width=1\linewidth]{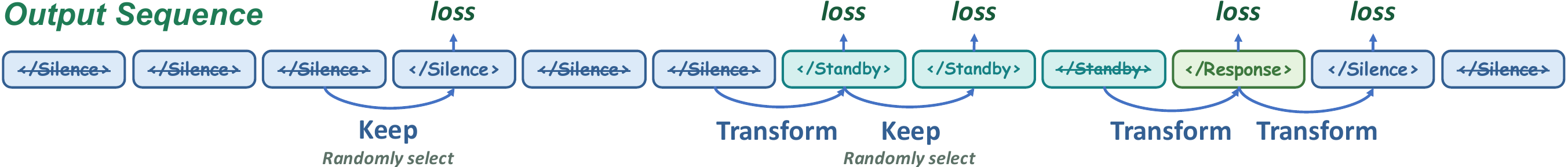}
  \caption{\textbf{State-transition supervision mask for streaming training.}
  The output sequence consists of \texttt{\textless/Silence\textgreater}, \texttt{\textless/Standby\textgreater}, and
  \texttt{\textless/Response\textgreater}. Blue arrows indicate the state-token positions whose losses are retained. We keep all \emph{Transform} positions, where the target state changes between adjacent windows, because they define the temporal decision boundaries. From the remaining continuation positions, we randomly select an equal number of \emph{Keep} positions to provide supervision for maintaining the current state.}
  \label{fig:method1-transition-mask}
\end{figure}

A direct causal-language-modeling loss over all state tokens is poorly balanced: long streams contain many more repeated \texttt{\textless/Silence\textgreater} tokens than state changes, causing the model to learn a conservative policy that rarely enters Standby or Response. Conversely, keeping losses only on state transitions creates a shortcut: the model can predict the next state from the previous state pattern without using visual evidence. We address both issues with \emph{state transition masking}, as shown in \Cref{fig:method1-transition-mask}.

Let \(s_t\) be the target state at streaming step \(t\). We always keep the loss for transition positions \(\mathcal{T}=\{t \mid s_t \ne s_{t-1}\}\), since these positions define the temporal decision boundaries. From the remaining continuation positions \(\mathcal{C}=\{t \mid s_t = s_{t-1}\}\), we uniformly sample a subset \(\widetilde{\mathcal{C}}\) with \(|\widetilde{\mathcal{C}}|=|\mathcal{T}|\). The masked state loss is
\[
\mathcal{L}_{\mathrm{state}}
= -\frac{1}{\sum_t m_t}
\sum_t m_t
\log p_{\theta}(s_t \mid V_{\le t}, y_{<t}),
\qquad
m_t = \mathbb{1}[t \in \mathcal{T} \cup \widetilde{\mathcal{C}}].
\]

This objective keeps every state switch while preserving an equal number of ``stay'' decisions, forcing the model to compare the previous state with the current visual evidence before deciding whether to maintain or change its behavior. In a typical streaming cycle, the effective supervision over state tokens follows the expected ratio
\[
\text{\texttt{\textless/Silence\textgreater}} :
\text{\texttt{\textless/Standby\textgreater}} :
\text{\texttt{\textless/Response\textgreater}}
= 2 : 2 : 1,
\]
because Silence and Standby each contain both transition and continuation supervision, whereas Response immediately hands control to answer generation. During training, the perception-window pixel budget is also teacher-forced from the previous target state; at inference time, the same controller is driven by the model's predicted state. This matches the training and deployment loops and lets the model learn both the temporal response policy and the active visual-budget policy in a single streaming instruction-tuning stage.

\paragraph{Data Mixture.}
Stage-3 focuses on streaming video interaction and long-video temporal reasoning. The training mixture contains 3.41M effective samples and 10.39M QAs, with key supervision coming from high-turn video data that requires incremental perception, response timing, temporal retrieval, and evidence-grounded reasoning over long contexts.

The streaming component is mainly provided by VideoChat3-OL, which contributes 191K video records but 4.69M trainable QA turns through action/event captioning, narration, QA, and explicit silence/standby turns. This teaches the model to accumulate evidence from incoming frames, decide when to respond, and avoid premature answers under incomplete visual context. The long-video component is centered on VideoChat3-LV, emphasizing event retrieval, timeline construction, temporal localization, subtitle- or narration-conditioned reasoning, and multi-event aggregation over extended videos. VideoChat3-Academic further provides Qwen3-VL-rewritten captions and QA across multiple duration buckets, improving the quality and coverage of both open-ended and multiple-choice video supervision.

Other sources serve as auxiliary supervision. Honey-Data~\cite{Honey-Data} is sampled at a low ratio to preserve broad image-language skills, while Dolci~\cite{olmo2025olmo3} stabilizes language quality. Additional video datasets, including TimeLens-100K~\cite{timelens}, Molmo2~\cite{molmo2}, Vript~\cite{Vript}, LLaVA-Video~\cite{llava_video}, CinePile~\cite{Cinepile}, MotionBench~\cite{motionbench}, TGIF/TVQA-style QA~\cite{TGIF-QA, TVQA}, AVA/EgoQA~\cite{AVA,egoqa}, and StreamForest~\cite{odvbench-streamforest}, provide complementary coverage of subtitle QA, dense captioning, motion understanding, tracking, spatial grounding, and driving-scene reasoning. In addition, following prior work on synthetic temporal primitives~\cite{jiang2026learning}, we synthesize a small set of atomic-motion questions, covering counting, rotation, and other basic motion skills in both single-choice and multiple-choice formats. Overall, Stage-3 prioritizes streaming behavior and long-context video reasoning while retaining general multimodal competence through carefully sampled auxiliary data.
\section{Evaluation}
\label{sec:experiment}

\subsection{General Video Understanding}
We evaluate VideoChat3 across four complementary dimensions of general video understanding: temporal perception, long-video understanding, complex video reasoning, and temporal grounding. Temporal perception is assessed on TOMATO~\cite{tomato}, MotionBench~\cite{motionbench}, TVBench~\cite{tvbench}, and TempCompass~\cite{tempcompass}, while long-video understanding is evaluated on Video-MME~\cite{videomme}, LVBench~\cite{lvbench}, and VideoEval-Pro~\cite{videoeval-pro}. For reasoning-intensive understanding, we use Video-MMMU~\cite{Video-mmmu}, MMVU~\cite{mmvu}, Minerva~\cite{minerva}, and Video-MME-v2~\cite{Video-MME-v2}. We further evaluate temporal grounding on the TimeLens suite~\cite{timelens}, covering Charades-STA~\cite{Charades-STA}, ActivityNet Captions~\cite{ActivityNet-Captions}, and QVHighlights~\cite{QVHighlights}, as well as VUE-TR v1~\cite{vidi}, VUE-TR v2~\cite{vidi2}, and the text-query subset of MomentSeeker~\cite{momentseeker}.

As shown in~\Cref{tab:video_benchmark_results}, VideoChat3-4B achieves strong and balanced performance among fully open video MLLMs. It obtains the best fully open results on MotionBench and TempCompass, scoring 61.7 and 75.6, respectively, demonstrating its ability to capture fine-grained motion and temporal dynamics. Compared with the open-weight Qwen3-VL-4B, VideoChat3 improves on \textbf{18 of the 19} directly comparable metrics, with the only decrease occurring on the open-ended split of VideoEval-Pro. Particularly large gains are observed on MMVU (+5.9), the three TimeLens splits (+9.7/+6.4/+8.3), VUE-TR v1/v2 (+15.0/+20.6), and MomentSeeker (+12.1).

Compared with similarly sized fully open models, VideoChat3 matches or outperforms Molmo2-4B on most benchmarks, including MotionBench, TempCompass, Video-MME, LVBench, VideoEval-Pro MCQ, Video-MMMU, MMVU, Minerva, Video-MME-v2, all three TimeLens splits, both VUE-TR versions, and MomentSeeker. Its gains over VideoChat-Flash-7B are especially pronounced on temporal grounding, reaching +16.4/+29.8/+34.3 on TimeLens, +30.6/+30.1 on VUE-TR v1/v2, and +18.7 on MomentSeeker. VideoChat3 also surpasses GPT-5 and Gemini 2.5 Flash on all three TimeLens splits, while remaining competitive with Gemini 2.5 Pro. Overall, these results highlight its ability to aggregate temporally ordered evidence and localize precise temporal boundaries.
 
\newcommand{\mycell}[2]{%
  \rotatebox{90}{%
    \parbox{2cm}{%
      \setlength{\baselineskip}{0.5em}%
      \textbf{\scriptsize{#1}}\\
      \footnotesize{\textcolor{gray}{#2}}%
    }%
  }%
}
\newcommand{\mycellthree}[3]{%
  \rotatebox{90}{%
    \parbox{2cm}{%
      \setlength{\baselineskip}{0.5em}%
      \textbf{\scriptsize{#1}}\\
      \footnotesize{\textcolor{gray}{#2}}\\
      \footnotesize{\textcolor{gray}{#3}}%
    }%
  }%
}
\newcommand{\mycellfour}[4]{%
  \rotatebox{90}{%
    \parbox{2cm}{%
      \setlength{\baselineskip}{0.5em}%
      {\textbf{\scriptsize{#1}}\textsuperscript{\fontsize{5.2pt}{6.2pt}\selectfont\textnormal{\textcolor{gray}{\,TL}}}}\\[0.25em]
      {\textbf{\scriptsize{#2}}\textsuperscript{\fontsize{5.2pt}{6.2pt}\selectfont\textnormal{\textcolor{gray}{\,TL}}}}\\[0.25em]
      {\textbf{\scriptsize{#3}}\textsuperscript{\fontsize{5.2pt}{6.2pt}\selectfont\textnormal{\textcolor{gray}{\,TL}}}}\\[0.25em]
      \footnotesize{\textcolor{gray}{#4}}%
    }%
  }%
}
\newcommand{\mycelltlsplit}[2]{%
  \rotatebox{90}{%
    \parbox{2cm}{%
      \setlength{\baselineskip}{0.55em}%
      {\textbf{\scriptsize{#1}}\textsuperscript{\fontsize{5.2pt}{6.2pt}\selectfont\textnormal{\textcolor{gray}{\,TL}}}}\\
      \footnotesize{\textcolor{gray}{#2}}%
    }%
  }%
}
\newcommand{\newcell}[1]{%
  \rotatebox{90}{%
    \parbox{2cm}{%
      \setlength{\baselineskip}{0.5em}%
      \textbf{\scriptsize{#1}}
    }%
  }%
}
\newcommand{\VCBenchNew}{\textemdash}
\colorlet{vc3tableLight}{vc3background!40!white}
\colorlet{vc3tableMid}{vc3background}
\colorlet{vc3tableDeep}{vc3accent!18!white}
\colorlet{molmocolor}{vc3accent}

\begin{table*}[t]
    \renewcommand{\arraystretch}{0.98}
    \centering
    \setlength{\tabcolsep}{2.0pt}
    \caption{\textbf{Video benchmark results} across temporal perception, long-video understanding, reasoning, and temporal grounding tasks. We compare VideoChat3-4B with proprietary, open-weight-only, and fully open baselines under the official metric of each benchmark, where higher scores indicate better performance. Temporal-grounding columns marked with `TL' share the TimeLens suite~\cite{timelens}. Bold and underlined numbers denote the best and second-best results among non-proprietary models, respectively.}
    \resizebox{\textwidth}{!}{
    \begin{tabular}{@{}l
    >{\columncolor{vc3tableLight}}c
    >{\columncolor{vc3tableLight}}c
    >{\columncolor{vc3tableLight}}c
    >{\columncolor{vc3tableLight}}c
    >{\columncolor{vc3tableMid}}c
    >{\columncolor{vc3tableMid}}c
    >{\columncolor{vc3tableMid}}c
    >{\columncolor{vc3tableMid}}c
    >{\columncolor{vc3tableLight}}c
    >{\columncolor{vc3tableLight}}c
    >{\columncolor{vc3tableLight}}c
    >{\columncolor{vc3tableLight}}c
    >{\columncolor{vc3tableLight}}c
    >{\columncolor{vc3tableMid}}c
    >{\columncolor{vc3tableMid}}c
    >{\columncolor{vc3tableMid}}c
    >{\columncolor{vc3tableMid}}c
    >{\columncolor{vc3tableMid}}c
    >{\columncolor{vc3tableMid}}c@{}}
        \multirow{2}{*}[-3pt]{\textbf{Model}}
        & \multicolumn{4}{c}{\scriptsize\textbf{Temporal Perception}}
        & \multicolumn{4}{c}{\scriptsize\textbf{Long Video}}
        & \multicolumn{5}{c}{\scriptsize\textbf{Reasoning}}
        & \multicolumn{6}{c}{\scriptsize\textbf{Temporal Grounding}} \\
        \cmidrule(lr){2-5}\cmidrule(lr){6-9}\cmidrule(lr){10-14}\cmidrule(lr){15-20}
        &
        \mycell{TOMATO}{test~\cite{tomato}}
        & \mycell{MotionBench}{val~\cite{motionbench}}
        & \mycell{TVBench}{test~\cite{tvbench}}
        & \mycell{TempCompass}{test MCQ~\cite{tempcompass}}
        & \mycell{Video-MME}{wo sub~\cite{videomme}}
        & \mycell{LVBench}{test~\cite{lvbench}}
        & \mycell{VideoEval-Pro}{MCQ~\cite{videoeval-pro}}
        & \mycell{VideoEval-Pro}{Open~\cite{videoeval-pro}}
        & \mycell{Video-MMMU}{test~\cite{Video-mmmu}}
        & \mycell{MMVU}{overall~\cite{mmvu}}
        & \mycell{Minerva}{test~\cite{minerva}}
        & \mycell{Video-MME-v2}{Non-Lin~\cite{Video-MME-v2}}
        & \mycell{Video-MME-v2}{Avg~\cite{Video-MME-v2}}
        & \mycelltlsplit{Charades}{mIoU~\cite{timelens}}
        & \mycelltlsplit{ActivityNet}{mIoU~\cite{timelens}}
        & \mycelltlsplit{QVHighlights}{mIoU~\cite{timelens}}
        & \mycell{VUE-TR V1}{mIOU~\cite{vidi}}
        & \mycell{VUE-TR V2}{mIOU~\cite{vidi2}}
        & \mycell{MomentSeeker}{mIoU~\cite{momentseeker}}\\
        \midrule
        \multicolumn{20}{@{}l}{\textbf{\textit{Proprietary Models}}} \\
        \textcolor{gray}{GPT-5~\cite{gpt5}}
        & \textcolor{gray}{53.0} & \textcolor{gray}{65.4} & \textcolor{gray}{\VCBenchNew} & \textcolor{gray}{80.4}
        & \textcolor{gray}{83.3} & \textcolor{gray}{65.2} & \textcolor{gray}{68.8} & \textcolor{gray}{\VCBenchNew}
        & \textcolor{gray}{84.6} & \textcolor{gray}{\VCBenchNew} & \textcolor{gray}{\VCBenchNew} & \textcolor{gray}{26.4} & \textcolor{gray}{44.7}
        & \textcolor{gray}{40.5} & \textcolor{gray}{42.9} & \textcolor{gray}{52.1} & \textcolor{gray}{\VCBenchNew} & \textcolor{gray}{20.0} & \textcolor{gray}{\VCBenchNew}\\
        \textcolor{gray}{GPT-5 mini~\cite{gpt5}}
        & \textcolor{gray}{44.1} & \textcolor{gray}{59.9} & \textcolor{gray}{\VCBenchNew} & \textcolor{gray}{74.9}
        & \textcolor{gray}{77.3} & \textcolor{gray}{54.7} & \textcolor{gray}{60.1} & \textcolor{gray}{\VCBenchNew}
        & \textcolor{gray}{82.5} & \textcolor{gray}{\VCBenchNew} & \textcolor{gray}{\VCBenchNew} & \textcolor{gray}{\VCBenchNew} & \textcolor{gray}{\VCBenchNew}
        & \textcolor{gray}{\VCBenchNew} & \textcolor{gray}{\VCBenchNew} & \textcolor{gray}{\VCBenchNew} & \textcolor{gray}{\VCBenchNew} & \textcolor{gray}{\VCBenchNew} & \textcolor{gray}{\VCBenchNew}\\
        \textcolor{gray}{Gemini 3 Pro~\cite{gemini3}}
        & \textcolor{gray}{48.3} & \textcolor{gray}{62.6} & \textcolor{gray}{\VCBenchNew} & \textcolor{gray}{82.8}
        & \textcolor{gray}{88.6} & \textcolor{gray}{77.0} & \textcolor{gray}{78.0} & \textcolor{gray}{\VCBenchNew}
        & \textcolor{gray}{87.6} & \textcolor{gray}{\VCBenchNew} & \textcolor{gray}{\VCBenchNew} & \textcolor{gray}{38.2} & \textcolor{gray}{56.8}
        & \textcolor{gray}{\VCBenchNew} & \textcolor{gray}{\VCBenchNew} & \textcolor{gray}{\VCBenchNew} & \textcolor{gray}{\VCBenchNew} & \textcolor{gray}{39.7} & \textcolor{gray}{\VCBenchNew}\\
        \textcolor{gray}{Gemini 2.5 Pro~\cite{comanici2025gemini}}
        & \textcolor{gray}{48.6} & \textcolor{gray}{62.0} & \textcolor{gray}{62.6} & \textcolor{gray}{81.9}
        & \textcolor{gray}{87.8} & \textcolor{gray}{75.7} & \textcolor{gray}{78.4} & \textcolor{gray}{44.2}
        & \textcolor{gray}{83.6} & \textcolor{gray}{\VCBenchNew} & \textcolor{gray}{\VCBenchNew} & \textcolor{gray}{\VCBenchNew} & \textcolor{gray}{\VCBenchNew}
        & \textcolor{gray}{52.8} & \textcolor{gray}{58.1} & \textcolor{gray}{70.4} & \textcolor{gray}{21.9} & \textcolor{gray}{\VCBenchNew} & \textcolor{gray}{\VCBenchNew}\\
        \textcolor{gray}{Gemini 2.5 Flash~\cite{comanici2025gemini}}
        & \textcolor{gray}{39.1} & \textcolor{gray}{59.3} & \textcolor{gray}{\VCBenchNew} & \textcolor{gray}{80.2}
        & \textcolor{gray}{84.2} & \textcolor{gray}{64.9} & \textcolor{gray}{69.6} & \textcolor{gray}{36.3}
        & \textcolor{gray}{\VCBenchNew} & \textcolor{gray}{\VCBenchNew} & \textcolor{gray}{\VCBenchNew} & \textcolor{gray}{\VCBenchNew} & \textcolor{gray}{\VCBenchNew}
        & \textcolor{gray}{48.6} & \textcolor{gray}{52.5} & \textcolor{gray}{64.3} & \textcolor{gray}{\VCBenchNew} & \textcolor{gray}{\VCBenchNew} & \textcolor{gray}{\VCBenchNew}\\
        \textcolor{gray}{Claude Sonnet 4.5~\cite{anthropic2025sonnet}}
        & \textcolor{gray}{39.6} & \textcolor{gray}{58.5} & \textcolor{gray}{\VCBenchNew} & \textcolor{gray}{72.8}
        & \textcolor{gray}{74.2} & \textcolor{gray}{50.5} & \textcolor{gray}{50.5} & \textcolor{gray}{\VCBenchNew}
        & \textcolor{gray}{\VCBenchNew} & \textcolor{gray}{\VCBenchNew} & \textcolor{gray}{\VCBenchNew} & \textcolor{gray}{\VCBenchNew} & \textcolor{gray}{\VCBenchNew}
        & \textcolor{gray}{\VCBenchNew} & \textcolor{gray}{\VCBenchNew} & \textcolor{gray}{\VCBenchNew} & \textcolor{gray}{\VCBenchNew} & \textcolor{gray}{\VCBenchNew} & \textcolor{gray}{\VCBenchNew}\\
        \midrule
        \multicolumn{20}{@{}l}{\textbf{\textit{Open weights only}}} \\
        MiniCPM-V-4.5-8B~\cite{minicpm-v4_5}
        & 29.8 & 59.7 & 53.0 & 72.7
        & 67.9 & 50.4 & 55.3 & 34.2
        & 57.1 & \textbf{58.9} & 36.0 & 11.8 & 27.1
        & 31.9 & 32.3 & 46.1 & 25.9 & 16.3 & 9.3\\
        Eagle2.5-8B~\cite{eagle2_5}
        & 31.0 & 55.7 & \underline{62.0} & \underline{74.4}
        & \textbf{72.4} & 50.9 & \underline{60.4} & 38.0
        & 48.3 & 54.3 & 37.6 & \underline{13.0} & \textbf{29.6}
        & \underline{53.6} & \underline{53.7} & \underline{64.0} & 30.2 & 8.7 & 5.3\\
        InternVideo2.5-8B~\cite{internvideo2_5}
        & 32.9 & 60.8 & 58.1 & 70.1
        & 65.1 & 46.4 & 53.2 & 27.2
        & 44.0 & 48.0 & \underline{37.7} & 11.2 & 26.3
        & 30.7 & 22.2 & 32.9 & 17.7 & 11.1 & 6.4\\
        VideoLLaMA3-7B~\cite{videolla3}
        & 26.8 & 48.7 & 49.5 & 68.1
        & 66.2 & 45.3 & 43.8 & 25.7
        & 34.6 & 39.9 & 29.5 & 10.9 & 25.7
        & 39.8 & 29.8 & 36.9 & 15.7 & 9.5 & 4.6\\
        Video-XL2-8B~\cite{videoxl2}
        & 28.4 & 54.3 & 47.5 & 58.5
        & 66.6 & 48.4 & 47.2 & 25.1
        & 39.9 & 50.0 & 35.3 & 7.3 & 20.0
        & 38.9 & 30.0 & 46.2 & 28.0 & 19.2 & 11.3\\
        InternVL3.5-4B~\cite{wang2025internvl3}
        & 26.8 & 56.5 & 57.1 & 68.8
        & 65.4 & 43.2 & 48.3 & 25.7
        & \textbf{57.6} & 47.6 & 31.3 & 9.9 & 24.6
        & 16.0 & 14.9 & 17.7 & 16.5 & 8.6 & 3.2\\
        Qwen3-VL-4B~\cite{qwen3vl}
        & 31.8 & 58.6 & 53.7 & 70.8
        & 69.3 & \underline{56.2} & 58.4 & \textbf{39.6}
        & 56.2 & 50.5 & 36.6 & 11.6 & 26.2
        & 46.4 & 48.2 & 58.7 & 32.9 & 19.6 & 13.8\\
        \midrule
        \multicolumn{20}{@{}l}{\textbf{\textit{Fully open models}}} \\
        PLM-8B~\cite{cho2025PerceptionLM}
        & 33.2 & 61.4 & 57.6 & 72.7
        & 58.3 & 44.5 & 47.2 & 24.3
        & 34.4 & 43.2 & 32.6 & 9.9 & 25.0
        & 34.5 & 7.6 & 4.2 & 5.9 & 3.0 & 2.5\\
        LLaVA-Video-7B~\cite{llava_video}
        & 24.9 & 54.2 & 51.1 & 66.6
        & 63.3 & 44.2 & 47.8 & 24.2
        & 36.1 & 47.1 & 35.3 & 7.2 & 19.9
        & 15.2 & 14.6 & 10.4 & 12.2 & 7.3 & 5.9\\
        VideoChat-Flash-7B~\cite{videochat-flash}
        & 32.5 & 60.6 & 54.4 & 69.4
        & 65.3 & 48.2 & 51.2 & 27.0
        & 40.0 & 47.5 & \textbf{40.1} & 9.9 & 25.0
        & 39.7 & 24.8 & 32.7 & 17.3 & 10.1 & 7.2\\
        Molmo2-4B~\cite{molmo2}
        & \textbf{39.8} & \underline{61.6} & \textbf{65.2} & 72.8
        & 69.6 & 53.9 & 59.9 & \underline{38.1}
        & 50.7 & 51.2 & 36.5 & 12.6 & \underline{29.0}
        & 33.3 & 39.8 & 58.7 & \underline{46.0} & \underline{32.7} & \underline{19.9}\\
        \textcolor{molmocolor}{\textbf{VideoChat3-4B}}
        & \underline{37.9} & \textbf{61.7} & 55.4 & \textbf{75.6}
        & \underline{70.1} & \textbf{56.7} & \textbf{60.7} & 37.8
        & \underline{57.4} & \underline{56.4} & 36.7 & \textbf{13.1} & 28.7
        & \textbf{56.1} & \textbf{54.6} & \textbf{67.0} & \textbf{47.9} & \textbf{40.2} & \textbf{25.9} \\
        \bottomrule
    \end{tabular}
        }%
    \label{tab:video_benchmark_results}
\end{table*}

\subsection{Streaming Video Understanding}

To further assess VideoChat3 in real-world streaming scenarios, we evaluate its online video understanding capability, where models must process visual inputs causally and continually update their internal state as new frames arrive. Unlike offline evaluation, which assumes access to the complete video, online evaluation emphasizes two complementary capabilities: maintaining an accurate memory of previously observed content and responding proactively at appropriate moments. We evaluate online perception and long-term memory on ODVBench~\cite{odvbench-streamforest}, OVBench~\cite{ovbench-videochatonline}, OVOBench~\cite{ovobench}, StreamingBench~\cite{streamingbench}, and River~\cite{river}, which collectively cover object-centric perception, event tracking, temporal state updates, and question answering over partially observed videos. We further evaluate proactive response ability on OVO-Timing~\cite{ovotiming-emgarde} and ProactiveVideoQA~\cite{proactivevideoqa}, which require models to identify response-worthy moments and generate timely, contextually appropriate answers during video streaming. As summarized in~\Cref{tab:online_video_benchmark_results}, we compare VideoChat3 with representative streaming and proactive video models, including VideoLLM-Online~\cite{videollmonline}, Flash-VStream~\cite{flashvstream}, Dispider~\cite{dispider}, VideoChat-Online~\cite{ovbench-videochatonline}, LiveCC~\cite{livecc}, TimeChat-Online~\cite{timechat-online}, StreamForest~\cite{odvbench-streamforest}, StreamingVLM~\cite{xu2025streamingvlm}, Streamo~\cite{streamo}, MMDuet-2~\cite{mmduet2}, and Em-Garde~\cite{ovotiming-emgarde}, as well as the general-purpose Qwen3-VL-4B baseline~\cite{qwen3vl}.

As shown in~\Cref{tab:online_video_benchmark_results}, VideoChat3-4B demonstrates that its strong offline video understanding capabilities can be effectively extended to online streaming scenarios. It achieves the best results on four of the six aggregate perception-and-memory metrics, including ODVBench, the task-average metric of OVOBench, StreamingBench, and River. In particular, VideoChat3 reaches 72.3 on ODVBench, surpassing StreamForest by 12.4 points, and exceeds the strongest competing results by 0.9 points on the OVOBench task average, 2.8 points on StreamingBench, and 2.9 points on River. These results show that VideoChat3 is capable not only of understanding complete videos offline, but also of continuously perceiving, retaining, and reasoning over evidence as a video unfolds online.

For proactive response, VideoChat3-4B further demonstrates the ability to determine when sufficient information has been accumulated and actively respond during streaming video interaction. It achieves an average F1 score of 35.5 on OVO-Timing, outperforming the specialized Em-Garde model by 4.5 points despite using no auxiliary 2B module. Compared with Qwen3-VL-4B under the same evaluation settings, VideoChat3 improves on \textbf{9 of the 11} directly comparable metrics overall. Among the proactive-response metrics, it achieves substantial gains on OVO-Timing (+27.4) and the EGO (+7.3) and VAD (+9.3) subsets of ProactiveVQA. Compared with the specialized MMDuet-2 model on ProactiveVQA, VideoChat3 achieves a higher score on VAD (38.9 vs. 28.9) but remains behind on the other three subsets. Overall, these results indicate that VideoChat3 extends offline video understanding beyond passive recognition: it can continuously interpret online video content, preserve relevant context over time, and proactively decide when and how to respond.

\providecommand{\VCBenchNew}{\textcolor{gray!65}{\textemdash}}

\providecommand{\vcOnlineRightOverhang}{2pt}

\newcommand{\onlinecell}[2]{%
  \rotatebox{60}{%
    \parbox{2.15cm}{%
      \centering
      \setlength{\baselineskip}{0.58em}%
      \textbf{\scriptsize #1}\\[2pt]
      {\footnotesize\textcolor{gray!75}{#2}}%
    }%
  }%
}

\newcommand{\pvqasubset}[1]{%
  \rotatebox{60}{%
    \textbf{\scriptsize #1}%
  }%
}

\newcommand{\vcmodel}[1]{%
  \textcolor{vc3OnlineModel}{\textbf{#1}}%
}

\newcommand{\addsize}[1]{%
  \textcolor{gray!60}{#1}%
}

\colorlet{vc3OnlineLight}{vc3background!28!white}
\colorlet{vc3OnlineMid}{vc3background!52!white}
\colorlet{vc3OnlineGroup}{vc3accent!7!white}
\colorlet{vc3OnlineModel}{vc3accent}

\begin{table*}[t]
    \centering
    \caption{\textbf{Online streaming video benchmark results.}
        Perception\&Memory benchmarks evaluate online video perception and memory,
        while proactive-response benchmarks evaluate timely and active response
        ability in streaming scenarios.
        The gray suffixes (e.g., \addsize{+1.5B}) denote the size of auxiliary modules.}
    \label{tab:online_video_benchmark_results}

    \vspace{2pt}
    \renewcommand{\arraystretch}{1.08}
    \setlength{\tabcolsep}{5pt}

    \resizebox{1.0\textwidth}{!}{%
    \begin{tabular}{@{}l
        c
        >{\columncolor{vc3OnlineLight}}c
        >{\columncolor{vc3OnlineLight}}c
        >{\columncolor{vc3OnlineLight}}c
        >{\columncolor{vc3OnlineLight}}c
        >{\columncolor{vc3OnlineLight}}c
        >{\columncolor{vc3OnlineLight}}c
        >{\columncolor{vc3OnlineMid}}c
        >{\columncolor{vc3OnlineMid}}c
        >{\columncolor{vc3OnlineMid}}c
        >{\columncolor{vc3OnlineMid}}c
        >{\columncolor{vc3OnlineMid}%
            [\tabcolsep][\vcOnlineRightOverhang]}c@{}}

        \textbf{Model}
        & \textbf{Size}
        & \multicolumn{6}{c}{%
            \textbf{\scriptsize Perception\&Memory}}
        & \multicolumn{5}{c}{%
            \textbf{\scriptsize Proactive Response}} \\
        \cmidrule(lr){3-8}
        \cmidrule(l){9-13}

        & &
        \onlinecell{OVBench}{%
            Avg.~\cite{ovbench-videochatonline}}
        & \onlinecell{ODVBench}{%
            Overall~\cite{odvbench-streamforest}}
        & \onlinecell{OVOBench}{%
            Overall~\cite{ovobench}}
        & \onlinecell{OVOBench}{%
            task Avg.~\cite{ovobench}}
        & \onlinecell{StreamingBench}{%
            Real-Time~\cite{streamingbench}}
        & \onlinecell{River}{%
            Avg.~\cite{river}}
        & \onlinecell{OVO-Timing}{%
            Avg. F1~\cite{ovotiming-emgarde}}
        & \multicolumn{4}{%
            >{\columncolor{vc3OnlineMid}%
                [\tabcolsep][\vcOnlineRightOverhang]}c@{}%
          }{%
            \onlinecell{ProactiveVQA}{\cite{proactivevideoqa}}%
          } \\[-1pt]

        & & & & & & & & &
        \pvqasubset{WEB}
        & \pvqasubset{EGO}
        & \pvqasubset{TV}
        & \pvqasubset{VAD} \\[3pt]

        \midrule

        \multicolumn{13}{%
            @{}>{\columncolor{vc3OnlineGroup}%
                [0pt][\vcOnlineRightOverhang]}l@{}%
          }{%
            \textbf{\textit{Online video methods}}%
          } \\

        VideoLLM-Online~\cite{videollmonline}
        & 7B
        & 9.6
        & \VCBenchNew
        & 12.8
        & 14.8
        & 36.0
        & \VCBenchNew
        & 6.9
        & 25.9   
        & 25.0   
        & 18.3   
        & 25.0   
        \\

        Flash-VStream~\cite{flashvstream}
        & 7B
        & 31.2
        & 35.7
        & 33.2
        & 32.3
        & 23.2
        & \VCBenchNew
        & 4.8
        & \VCBenchNew   
        & \VCBenchNew   
        & \VCBenchNew   
        & \VCBenchNew   
        \\

        Dispider~\cite{dispider}
        & 7B\addsize{+1.5B}
        & \VCBenchNew
        & 45.2
        & 41.8
        & 45.0
        & 67.6
        & \VCBenchNew
        & \VCBenchNew
        & \VCBenchNew   
        & \VCBenchNew   
        & \VCBenchNew   
        & \VCBenchNew   
        \\

        VideoChat-Online~\cite{ovbench-videochatonline}
        & 4B
        & 54.9
        & 54.5
        & \VCBenchNew
        & \VCBenchNew
        & \VCBenchNew
        & \VCBenchNew
        & \VCBenchNew
        & \VCBenchNew   
        & \VCBenchNew   
        & \VCBenchNew   
        & \VCBenchNew   
        \\

        LiveCC~\cite{livecc}
        & 7B
        & 46.7
        & 44.8
        & \textbf{59.8}
        & \VCBenchNew
        & 71.4
        & 33.9
        & \VCBenchNew
        & \VCBenchNew   
        & \VCBenchNew   
        & \VCBenchNew   
        & \VCBenchNew   
        \\

        TimeChat-Online~\cite{timechat-online}
        & 7B
        & \VCBenchNew
        & \VCBenchNew
        & 47.6
        & 51.0
        & 75.4
        & \VCBenchNew
        & \VCBenchNew
        & \VCBenchNew   
        & \VCBenchNew   
        & \VCBenchNew   
        & \VCBenchNew   
        \\


        StreamForest~\cite{odvbench-streamforest}
        & 7B
        & \textbf{60.5}
        & 59.9
        & 55.6
        & 57.0
        & 77.3
        & \VCBenchNew
        & 14.0
        & \VCBenchNew   
        & \VCBenchNew   
        & \VCBenchNew   
        & \VCBenchNew   
        \\

        StreamingVLM~\cite{xu2025streamingvlm}
        & 7B
        & 47.8
        & 52.5
        & \VCBenchNew
        & \VCBenchNew
        & 73.2
        & 39.9
        & \VCBenchNew
        & \VCBenchNew   
        & \VCBenchNew   
        & \VCBenchNew   
        & \VCBenchNew   
        \\

        Streamo~\cite{streamo}
        & 7B
        & \VCBenchNew
        & \VCBenchNew
        & 57.9
        & 60.3
        & \VCBenchNew
        & \VCBenchNew
        & \VCBenchNew
        & \VCBenchNew   
        & \VCBenchNew   
        & \VCBenchNew   
        & \VCBenchNew   
        \\

        MMDuet-2~\cite{mmduet2}
        & 3B
        & 47.1
        & 50.0
        & 48.8
        & 52.3
        & 71.5
        & 33.6
        & 20.5
        & \textbf{53.3}   
        & \textbf{33.6}   
        & \textbf{43.4}   
        & 28.9   
        \\

        Em-Garde~\cite{ovotiming-emgarde}
        & 7B\addsize{+2B}
        & 47.2
        & 47.7
        & 50.3
        & \VCBenchNew
        & 76.7
        & 33.4
        & 31.0
        & \VCBenchNew   
        & \VCBenchNew   
        & \VCBenchNew   
        & \VCBenchNew   
        \\

        Qwen3-VL-4B~\cite{qwen3vl}
        & 4B
        & 54.5
        & 57.4
        & 56.7
        & 61.6
        & 80.2
        & 37.8
        & 8.1
        & 42.4   
        & 25.0   
        & 40.4   
        & 29.6   
        \\

        \midrule

        \vcmodel{VideoChat3-4B}
        & 4B
        & 60.0
        & \textbf{72.3}
        & 57.8
        & \textbf{62.5}
        & \textbf{83.0}
        & \textbf{42.8}
        & \textbf{35.5}
        & 39.6   
        & 32.3   
        & 39.6   
        & \textbf{38.9}   
        \\

        \bottomrule
    \end{tabular}%
    }
\end{table*}

\subsection{Efficiency Analysis}

As shown in \Cref{tab:inference_cost}, VideoChat3 introduces explicit spatio-temporal modeling in the vision encoder, which leads to a higher encoder latency than Qwen3-VL. However, this design substantially reduces the number of visual tokens passed to the LLM. Under the controlled setting where both models use the same number of ViT patches per frame, VideoChat3 consistently produces only half as many visual tokens as Qwen3-VL. This token reduction is particularly important because the computational cost of the LLM grows quadratically with sequence length, whereas the additional cost in the vision encoder increases approximately linearly with the number of input frames.


\begin{table*}[t]
\centering
\caption{Inference cost comparison on NVIDIA H200 using the HuggingFace Pipeline with FlashAttention-2.}
\label{tab:inference_cost}
\small
\setlength{\tabcolsep}{4.5pt}
\renewcommand{\arraystretch}{1.12}
\sisetup{
  detect-weight=true,
  detect-family=true,
  table-number-alignment=center
}
\begin{tabular}{
  @{}l
  S[table-format=4.0]
  S[table-format=6.0]
  S[table-format=2.3]
  S[table-format=3.2]
  S[table-format=2.3]
  S[table-format=2.3]
  S[table-format=2.3]
  @{}
}
\toprule
\multirow{2}{*}{\textbf{Model}} &
{\multirow{2}{*}{\makecell{\textbf{Input}\\\textbf{Frames}}}} &
{\multirow{2}{*}{\makecell{\textbf{Visual}\\\textbf{Tokens}}}} &
{\multirow{2}{*}{\makecell{\textbf{FLOPs}\\\textbf{($\times 10^{15}$)}}}} &
{\multirow{2}{*}{\makecell{\textbf{GPU Mem.}\\\textbf{(GB)}}}} &
\multicolumn{3}{c@{}}{\textbf{Latency (s)}} \\
\cmidrule(l){6-8}
& & & & &
{\makecell{\textbf{Vision}\\\textbf{Encoder}}} &
{\textbf{LLM}} &
{\textbf{Total}} \\
\midrule
Qwen3-VL   & 256  & 25088  & 0.466 & 20.583  & 0.288  & 1.074  & 1.363  \\
\rowcolor{VideoChatGreen!70} VideoChat3 & 256  & 12544  & 0.385 & 17.663  & 1.272  & 0.380  & 1.652  \\
\addlinespace[2pt]
Qwen3-VL   & 512  & 50176  & 1.341 & 32.915  & 0.578  & 3.260  & 3.838  \\
\rowcolor{VideoChatGreen!70} VideoChat3 & 512  & 25088  & 0.864 & 26.676  & 2.595  & 1.001  & 3.596  \\
\addlinespace[2pt]
Qwen3-VL   & 1024 & 100352 & 4.313  & 57.581  & 1.152  & 11.098 & 12.251 \\
\rowcolor{VideoChatGreen!70} VideoChat3 & 1024 & 50176  & 2.108  & 44.709  & 5.158  & 2.941  & 8.099  \\
\addlinespace[2pt]
Qwen3-VL   & 2048 & 200704 & 15.150 & 106.913 & 2.313  & 42.135 & 44.449 \\
\rowcolor{VideoChatGreen!70} VideoChat3 & 2048 & 100352 & 5.738  & 80.775  & 10.297 & 10.115 & 20.412 \\
\bottomrule
\end{tabular}
\end{table*}

This trade-off becomes increasingly beneficial for long-video inference. For 512 input frames, VideoChat3 reduces FLOPs from 1.341 to 0.864 \(\times 10^{15}\), lowers GPU memory from 32.92 GB to 26.68 GB, and achieves a lower total latency than Qwen3-VL. The advantage becomes more pronounced as the video length increases: at 1024 frames, VideoChat3 reduces total latency from 12.251s to 8.099s, and at 2048 frames, it reduces latency from 44.449s to 20.412s, while also cutting FLOPs by more than 60\% and reducing memory usage by 26.14 GB. These results demonstrate that VideoChat3's early spatio-temporal compression effectively shifts computation from the quadratic LLM stage to the linear vision-encoding stage, making it substantially more efficient and scalable for long-video understanding.

\subsection{Ablation Studies}

\paragraph{I3D-ViT} As presented in \Cref{tab:image_video_vit_ablation}, we benchmark MoonViT~\cite{kimivl}---whose weights serve as the initialization parameters for our model---against our fully trained I3D-ViT across a variety of downstream Video MLLM tasks. As shown in \Cref{tab:image_vit_ablation}, compared with MoonViT, which has been thoroughly pre-trained on large-scale image-text corpora, I3D-ViT achieves comparable or even marginally superior performance. As observed from \Cref{tab:video_vit_ablation}, I3D-ViT consistently and significantly outperforms MoonViT across all categories of video tasks. We attribute this performance gain entirely to the frame sampling advantage of I3D-ViT: enabled by its 4$\times$ temporal compression capability, I3D-ViT supports nearly four times the number of sampled video frames relative to MoonViT in both training and inference phases.

\begin{table*}[!ht]
    \renewcommand{\arraystretch}{0.98}
    \centering
    \caption{\textbf{Comparison of MoonViT and I3D-ViT on different VLM Benchmarks.}}
    \label{tab:image_video_vit_ablation}

    \begin{subtable}{\textwidth}
    \centering
    \setlength{\tabcolsep}{3.8pt}
    \caption{\textbf{Comparison on Image Benchmarks.} Image Avg. normalizes OCRBench to a 100-point scale.}
    \label{tab:image_vit_ablation}
    \resizebox{0.80\textwidth}{!}{
    \begin{tabular}{@{}l
    >{\columncolor{vc3tableLight}}c>{\columncolor{vc3tableLight}}c>{\columncolor{vc3tableLight}}c>{\columncolor{vc3tableLight}}c
    >{\columncolor{vc3tableMid}}c>{\columncolor{vc3tableMid}}c>{\columncolor{vc3tableMid}}c>{\columncolor{vc3tableMid}}c
    >{\columncolor{vc3tableLight}}c>{\columncolor{vc3tableLight}}c>{\columncolor{vc3tableLight}}c
    >{\columncolor{vc3tableDeep}}c@{}}
        \multirow{2}{*}[-3pt]{\textbf{Visual Tokenizer}}
        & \multicolumn{4}{c}{\scriptsize\textbf{Image Perception}}
        & \multicolumn{4}{c}{\scriptsize\textbf{OCR \& Chart}}
        & \multicolumn{3}{c}{\scriptsize\textbf{Image STEM}}
        & \multicolumn{1}{c}{\scriptsize\textbf{Avg.}} \\
        \cmidrule(lr){2-5}\cmidrule(lr){6-9}\cmidrule(lr){10-12}\cmidrule(lr){13-13}
        & \mycell{MMBench}{DEV EN V1.1}
        & \mycell{RealWorldQA}{test}
        & \mycell{AI2D}{test w. M.}
        & \mycell{MMStar}{test}
        & \mycell{OCRBench}{score}
        & \mycell{ChartQA}{test}
        & \mycell{DocVQA}{val}
        & \mycell{InfoVQA}{val}
        & \mycell{MMMU}{dev val}
        & \mycell{MathVista}{mini}
        & \mycell{MathVision}{mini}
        & \newcell{Image Avg.} \\
        \midrule

        MoonViT~\cite{kimivl}
        & 74.5 & \textbf{68.9} & 74.8 & 57.7
        & 709 & \textbf{75.2} & 90.8 & 66.5
        & \textbf{53.1} & 65.8 & 20.1
        & 65.30 \\

        \textbf{I3D-ViT (ours)}
        & \textbf{74.9} & 67.7 & \textbf{75.5} & \textbf{60.6}
        & \textbf{743} & 74.3 & 90.8 & \textbf{69.1}
        & 52.0 & \textbf{66.8} & \textbf{21.1}
        & \textbf{66.10} \\

        \bottomrule
    \end{tabular}
    }
    \end{subtable}

    \vspace{0.7em}

    \begin{subtable}{\textwidth}
    \centering
    \setlength{\tabcolsep}{4.2pt}
    \caption{\textbf{Comparison on Video Benchmarks.} Temporal-grounding columns \textsuperscript{TL} share the TimeLens suite and differ by split.}
    \label{tab:video_vit_ablation}
    \resizebox{\textwidth}{!}{
    \begin{tabular}{@{}l
    >{\columncolor{vc3tableLight}}c>{\columncolor{vc3tableLight}}c>{\columncolor{vc3tableLight}}c>{\columncolor{vc3tableLight}}c
    >{\columncolor{vc3tableMid}}c>{\columncolor{vc3tableMid}}c>{\columncolor{vc3tableMid}}c
    >{\columncolor{vc3tableLight}}c>{\columncolor{vc3tableLight}}c>{\columncolor{vc3tableLight}}c
    >{\columncolor{vc3tableMid}}c>{\columncolor{vc3tableMid}}c>{\columncolor{vc3tableMid}}c
    >{\columncolor{vc3tableDeep}}c@{}}
        \multirow{2}{*}[-3pt]{\textbf{Visual Tokenizer}}
        & \multicolumn{4}{c}{\scriptsize\textbf{Video Perception}}
        & \multicolumn{3}{c}{\scriptsize\textbf{Long Video}}
        & \multicolumn{3}{c}{\scriptsize\textbf{Reasoning}}
        & \multicolumn{3}{c}{\scriptsize\textbf{Temporal Grounding}}
        & \multicolumn{1}{c}{\scriptsize\textbf{Avg.}} \\
        \cmidrule(lr){2-5}\cmidrule(lr){6-8}\cmidrule(lr){9-11}\cmidrule(lr){12-14}\cmidrule(lr){15-15}
        & \mycell{TOMATO}{test}
        & \mycell{MotionBench}{val}
        & \mycell{TVBench}{test}
        & \mycell{TempCompass}{test MCQ}
        & \mycell{Video-MME}{overall}
        & \mycell{LongVideoBench}{test}
        & \mycell{LVBench}{test}
        & \mycell{Video-MMMU}{test}
        & \mycell{MMVU}{overall}
        & \mycell{Minerva}{test}
        & \mycelltlsplit{Charades}{mIoU}
        & \mycelltlsplit{ActivityNet}{mIoU}
        & \mycelltlsplit{QVHighlights}{mIoU}
        & \newcell{Video Avg.} \\
        \midrule

        MoonViT~\cite{kimivl}
        & 24.5 & 53.3 & 47.7 & 70.2
        & 59.9 & 55.5 & 39.4
        & 46.9 & 51.0 & 30.9
        & 30.7 & 27.1 & 33.9
        & 43.9 \\

        \textbf{I3D-ViT (ours)}
        & \textbf{25.3} & \textbf{55.6} & \textbf{54.6} & \textbf{74.1}
        & \textbf{65.4} & \textbf{60.7} & \textbf{45.0}
        & \textbf{53.4} & \textbf{51.8} & \textbf{34.8}
        & \textbf{45.0} & \textbf{43.7} & \textbf{58.4}
        & \textbf{51.4} \\

        \bottomrule
    \end{tabular}
    }
    \end{subtable}
\end{table*}

\paragraph{Dynamic Stream and Training Mask} We isolate the two streaming-specific components on OVO-Timing~\cite{ovotiming-emgarde}, whose average F1 directly measures whether a model responds around the correct moment. First, we fix the proposed state-transition mask and compare three perception-window policies: always-low budget, always-high budget, and the state-conditioned dynamic budget used by VideoChat3. This separates the benefit of actively enlarging the next window after Standby from simply spending more visual tokens everywhere. Second, we fix the dynamic window policy and vary only the state-token supervision: loss on all state tokens, loss only on state transitions, and our balanced state-transition mask. This ablation tests whether the mask improves timing decisions by preserving transition supervision while preventing the model from being dominated by repeated Silence labels.

\providecommand{\VCBenchNew}{\textemdash}

\begin{table}[ht]
    \renewcommand{\arraystretch}{1.05}
    \centering
    \caption{\textbf{Ablation design for dynamic streaming perception and state-transition masking.} We evaluate all variants on OVO-Timing~\cite{ovotiming-emgarde}. The left table fixes the state-transition mask and compares perception-window policies; the right table fixes the dynamic perception window and compares state-token supervision strategies. The budget ratio is reported only for window-policy variants and is normalized by the always-high \(448{\times}448\) policy.}
    \begin{subtable}[t]{0.48\linewidth}
        \centering
        \setlength{\tabcolsep}{3pt}
        \resizebox{\linewidth}{!}{
        \begin{tabular}{@{}lcccr@{}} 
            \toprule
            \textbf{Window Policy}
            & \textbf{F1 \(\uparrow\)}
            & \textbf{P \(\uparrow\)}   
            & \textbf{R \(\uparrow\)}   
            & \textbf{Budget \(\downarrow\)} \\
            \midrule
            Fixed low \((224^2)\)
            & 33.5
            & 31.9 & 37.0 
            & 25\% \\
            Fixed high \((448^2)\)
            & 30.5
            & 22.3 & \textbf{58.8}
            & 100\% \\
            Dynamic \((224^2 \& 448^2)\)
            & \textbf{35.5}
            & \textbf{33.1} & 44.9 
            & 30.2\% \\
            \bottomrule
        \end{tabular}
        }
        \caption{Perception-window policy.}
    \end{subtable}
    \hfill
    \begin{subtable}[t]{0.39\linewidth}
        \centering
        \setlength{\tabcolsep}{4pt}
        \resizebox{\linewidth}{!}{
        \begin{tabular}{@{}lccc@{}} 
            \toprule
            \textbf{State Supervision}
            & \textbf{F1 \(\uparrow\)}
            & \textbf{P \(\uparrow\)}   
            & \textbf{R \(\uparrow\)} \\ 
            \midrule
            All state tokens
            & 5.8
            & 7.5 & 5.1 \\
            Transitions only
            & 20.1
            & 11.8 & \textbf{97.9} \\
            State-transition mask
            & \textbf{35.5}
            & \textbf{33.1} & 44.9 \\
            \bottomrule
        \end{tabular}
        }
        \caption{State-token supervision.}
    \end{subtable}
    \label{tab:ablation_dynamic_stream_mask}
\end{table}

\paragraph{VideoChat3-Academic2M Pipeline}
As shown in \Cref{tab:ablation_data_academic2m}, we benchmark model performance when trained with raw annotations against that when trained with annotations enhanced via our VideoChat3-Academic2M pipeline. Equipped with our integrated enhancement and error filtering pipeline, the model yields consistent improvements across nearly all video understanding capabilities, most notably temporal perception and temporal grounding. This finding validates the effectiveness of our Evidence-Grounded Annotation Enhancement method for boosting temporal modeling performance.

\begin{table*}[!ht]
    \renewcommand{\arraystretch}{0.98}
    \centering
    \setlength{\tabcolsep}{2.9pt}
    \caption{\textbf{Ablation study on VideoChat3-Academic2M pipeline.}}
    \resizebox{\textwidth}{!}{
    \begin{tabular}{@{}l
    >{\columncolor{vc3tableLight}}c
    >{\columncolor{vc3tableLight}}c
    >{\columncolor{vc3tableLight}}c
    >{\columncolor{vc3tableLight}}c
    >{\columncolor{vc3tableMid}}c
    >{\columncolor{vc3tableMid}}c
    >{\columncolor{vc3tableMid}}c
    >{\columncolor{vc3tableLight}}c
    >{\columncolor{vc3tableLight}}c
    >{\columncolor{vc3tableLight}}c
    >{\columncolor{vc3tableMid}}c
    >{\columncolor{vc3tableMid}}c
    >{\columncolor{vc3tableMid}}c
    >{\columncolor{vc3tableDeep}}c@{}}
        \multirow{2}{*}[-3pt]{\textbf{Data for Stage2}}
        & \multicolumn{4}{c}{\scriptsize\textbf{Temporal Perception}}
        & \multicolumn{3}{c}{\scriptsize\textbf{Long Video}}
        & \multicolumn{3}{c}{\scriptsize\textbf{Reasoning}}
        & \multicolumn{3}{c}{\scriptsize\textbf{Temporal Grounding}}
        & \multicolumn{1}{c}{\scriptsize\textbf{Avg.}} \\
        \cmidrule(lr){2-5}\cmidrule(lr){6-8}\cmidrule(lr){9-11}\cmidrule(lr){12-14}\cmidrule(lr){15-15}
        &
        \mycell{TOMATO}{test}
        & \mycell{MotionBench}{val}
        & \mycell{TVBench}{test}
        & \mycell{TempCompass}{test MCQ}
        & \mycell{Video-MME}{wo sub}
        & \mycell{LongVideoBench}{test}
        & \mycell{LVBench}{test}
        & \mycell{Video-MMMU}{test}
        & \mycell{MMVU}{overall}
        & \mycell{Minerva}{test}
        & \mycelltlsplit{Charades}{mIoU}
        & \mycelltlsplit{ActivityNet}{mIoU}
        & \mycelltlsplit{QVHighlights}{mIoU}
        & \newcell{Overall} \\
        \midrule

        Original Annotation
        & 24.7 & 54.4 & 44.0 & 67.8
        & 64.4 & 59.7 & \textbf{48.7}
        & 50.9 & \textbf{53.0} & 34.7
        & 38.1 & 24.6 & 29.5
        & 45.7 \\

        \textbf{ + VideoChat3-Academic2M}
        & \textbf{25.3} & \textbf{55.6} & \textbf{54.6} & \textbf{74.1}
        & \textbf{65.4} & \textbf{60.7} & 45.0
        & \textbf{53.4} & 51.8 & \textbf{34.8}
        & \textbf{45.0} & \textbf{43.7} & \textbf{58.4}
        & \textbf{51.4} \\
        
        \bottomrule
    \end{tabular}
}
    \label{tab:ablation_data_academic2m}
\end{table*}

\paragraph{VideoChat3-LV116K}

As shown in \Cref{tab:ablation_data_lv116k}, VideoChat3-LV116K significantly boosts the performance on long-video benchmarks including Video-MME, LongVideoBench~\cite{LongVideoBench}, and LVBench, which validates the effectiveness of our data construction pipeline. Meanwhile, it also yields substantial improvements on temporal grounding tasks for longer videos, such as ActivityNet and QVHighlights.

\begin{table*}[!ht]
    \renewcommand{\arraystretch}{0.9}
    \centering
    \setlength{\tabcolsep}{2.9pt}
    \caption{\textbf{Ablation study on VideoChat3-LV116K pipeline.} }
    \resizebox{\textwidth}{!}{
    \begin{tabular}{@{}l
    >{\columncolor{vc3tableLight}}c
    >{\columncolor{vc3tableLight}}c
    >{\columncolor{vc3tableLight}}c
    >{\columncolor{vc3tableLight}}c
    >{\columncolor{vc3tableMid}}c
    >{\columncolor{vc3tableMid}}c
    >{\columncolor{vc3tableMid}}c
    >{\columncolor{vc3tableLight}}c
    >{\columncolor{vc3tableLight}}c
    >{\columncolor{vc3tableLight}}c
    >{\columncolor{vc3tableMid}}c
    >{\columncolor{vc3tableMid}}c
    >{\columncolor{vc3tableMid}}c
    >{\columncolor{vc3tableDeep}}c@{}}
        \multirow{2}{*}[-3pt]{\textbf{Data for Stage3}}
        & \multicolumn{4}{c}{\scriptsize\textbf{Temporal Perception}}
        & \multicolumn{3}{c}{\scriptsize\textbf{Long Video}}
        & \multicolumn{3}{c}{\scriptsize\textbf{Reasoning}}
        & \multicolumn{3}{c}{\scriptsize\textbf{Temporal Grounding}}
        & \multicolumn{1}{c}{\scriptsize\textbf{Avg.}} \\
        \cmidrule(lr){2-5}\cmidrule(lr){6-8}\cmidrule(lr){9-11}\cmidrule(lr){12-14}\cmidrule(lr){15-15}
        &
        \mycell{TOMATO}{test}
        & \mycell{MotionBench}{val}
        & \mycell{TVBench}{test}
        & \mycell{TempCompass}{test MCQ}
        & \mycell{Video-MME}{wo sub}
        & \mycell{LongVideoBench}{test}
        & \mycell{LVBench}{test}
        & \mycell{Video-MMMU}{test}
        & \mycell{MMVU}{overall}
        & \mycell{Minerva}{test}
        & \mycelltlsplit{Charades}{mIoU}
        & \mycelltlsplit{ActivityNet}{mIoU}
        & \mycelltlsplit{QVHighlights}{mIoU}
        & \newcell{Overall} \\
        \midrule

        Baseline
        & 38.8 & 61.8 & \textbf{57.7} & 74.6
        & 68.4 & 63.6 & 54.3
        & 53.3 & 55.6 & \textbf{38.1}
        & 55.8 & 50.5 & 60.3
        & 56.4 \\

        \textbf{ + VideoChat3-LV116K}
        & 38.8 & \textbf{62.4} & 57.6 & \textbf{74.9}
        & \textbf{70.2} & \textbf{64.2} & \textbf{56.7}
        & \textbf{55.8} & \textbf{56.4} & 37.8
        & \textbf{55.8} & \textbf{54.8} & \textbf{67.7}
        & \textbf{57.9} \\
        
        \bottomrule
    \end{tabular}
}
    \label{tab:ablation_data_lv116k}
\end{table*}

\paragraph{VideoChat3-OL617K}
As shown in \Cref{tab:ablation_data_ol617k}, VideoChat3-OL617K improves both the content and timing of proactive responses, raising OVO-Timing Avg. F1 from 4.0 to 35.5. On ProactiveVQA, it improves EGO, TV, and VAD by 7.3, 5.3, and 11.5 points, respectively, while WEB decreases by 12.3 points. Because PAUC, ProactiveVQA's evaluation metric, does not explicitly penalize response frequency or exact duplicates, these scores may underrepresent repetitive or excessive responses; qualitatively, the trained model responds more naturally and appropriately. VideoChat3-OL617K also improves online perception and memory benchmarks while leaving offline video QA largely stable, demonstrating that our data construction pipeline provides effective supervision for proactive video understanding without compromising general competence.

\begin{table*}[!ht]
    \renewcommand{\arraystretch}{0.9}
    \centering
    \setlength{\tabcolsep}{2.9pt}

    \caption{\textbf{Ablation study on VideoChat3-OL617K pipeline.}}
    \label{tab:ablation_data_ol617k}

    \resizebox{\textwidth}{!}{%
    \begin{tabular}{@{}l
        >{\columncolor{vc3tableLight}}c
        >{\columncolor{vc3tableLight}}c
        >{\columncolor{vc3tableLight}}c
        >{\columncolor{vc3tableLight}}c
        >{\columncolor{vc3tableLight}}c
        >{\columncolor{vc3tableLight}}c
        >{\columncolor{vc3tableMid}}c
        >{\columncolor{vc3tableMid}}c
        >{\columncolor{vc3tableMid}}c
        >{\columncolor{vc3tableMid}}c
        >{\columncolor{vc3tableMid}}c
        >{\columncolor{vc3tableLight}}c
        >{\columncolor{vc3tableLight}}c
        >{\columncolor{vc3tableLight}}c
        >{\columncolor{vc3tableLight}}c@{}}

        \textbf{Data for Stage3}
        & \multicolumn{6}{c}{%
            \scriptsize\textbf{Perception\&Memory}}
        & \multicolumn{5}{c}{%
            \scriptsize\textbf{Proactive Response}}
        & \multicolumn{4}{c}{%
            \scriptsize\textbf{Offline Video QA}} \\

        \cmidrule(lr){2-7}
        \cmidrule(lr){8-12}
        \cmidrule(lr){13-16}

        &
        \mycell{OVBench}{Avg.}
        & \mycell{ODVBench}{Overall}
        & \mycell{OVOBench}{Overall}
        & \mycell{OVOBench}{task Avg.}
        & \mycell{StreamingBench}{Real-Time}
        & \mycell{River}{Avg.}
        & \mycell{OVO-Timing}{Avg. F1}
        & \multicolumn{4}{>{\columncolor{vc3tableMid}}c}{%
            \mycell{ProactiveVQA}{}}
        & \mycell{TOMATO}{test}
        & \mycell{Video-MME}{wo sub}
        & \mycell{Video-MMMU}{test}
        & \mycelltlsplit{Charades}{mIoU} \\[-1pt]

        & & & & & & & &
        \scriptsize WEB
        & \scriptsize EGO
        & \scriptsize TV
        & \scriptsize VAD
        & & & & \\[3pt]

        \midrule

        Original Annotation
        & 57.6
        & 58.2
        & 55.7
        & 60.5
        & 81.9
        & 38.3
        & 4.0
        & \textbf{51.9}  
        & 25.0  
        & 34.3  
        & 27.4  
        & \textbf{38.8}
        & \textbf{70.2}
        & 55.8
        & 55.8
        \\

        \textbf{+ VideoChat3-OL617K}
        & \textbf{60.0}
        & \textbf{72.3}
        & \textbf{57.8}
        & \textbf{62.5}
        & \textbf{83.0}
        & \textbf{42.8}
        & \textbf{35.5}
        & 39.6  
        & \textbf{32.3}
        & \textbf{39.6}
        & \textbf{38.9}
        & 37.8
        & 70.0
        & \textbf{56.7}
        & \textbf{56.1}
        \\

        \bottomrule
    \end{tabular}%
    }
\end{table*}

\section{Related Work}
\label{sec:related}

\paragraph{Video Multimodal Large Language Models.}
Recent video multimodal large language models (Video MLLMs) extend the image-MLLM paradigm to dynamic visual inputs by coupling a visual encoder, a projection module, and a large language model. Early systems such as VideoChat~\cite{li2023videochat}, Video-ChatGPT~\cite{maaz2023videochatgpt}, and Video-LLaVA~\cite{lin2023videollava} demonstrated that video-centric instruction tuning can enable open-ended video dialogue and temporal reasoning. Subsequent work further improved the data and training recipes: LLaVA-Video~\cite{llava_video} introduced large-scale synthetic video instruction data, LLaVA-OneVision~\cite{li2024llavaonevision} showed strong transfer across image, multi-image, and video scenarios, and VideoLLaMA3~\cite{videolla3} emphasized vision-centric training for unified image and video understanding. More recent models, including InternVideo2.5~\cite{internvideo2_5}, Qwen3-VL~\cite{qwen3vl}, Molmo2~\cite{molmo2}, and LLaVA-OneVision-2~\cite{an2026llavaov2}, have pushed open-weight video understanding toward stronger long-context reasoning, grounding, and broad multimodal capability, with adjacent VLM adaptation and retrieval studies exploring related transfer settings~\cite{zhu2024awt,zhu2024efficient,zhu2025freeret}. Despite this progress, many models still inherit a frame-sampling interface from image MLLMs, where videos are represented as a sparse set of independently encoded frames. This design is simple and effective for short clips, but it can discard fine-grained motion cues~\cite{zhang2025mog}, scale poorly to long videos, and offer limited support for causal streaming interaction. VideoChat3 instead treats video as a spatiotemporal signal from the visual tokenizer onward, aiming for a fully open, efficient, and generalist Video MLLM that covers short, long, grounded, and streaming video understanding.

\paragraph{Efficient Video Tokenization and Long-Video Modeling.}
The token explosion caused by dense video inputs has motivated a large body of work on efficient video representation. A common direction is to compress visual tokens after per-frame encoding, for example through hierarchical clip-to-video compression in VideoChat-Flash~\cite{videochat-flash}, adaptive spatiotemporal compression in LongVU~\cite{shen2024longvu}, and task-aware KV sparsification in Video-XL-2~\cite{videoxl2}. Another recent direction exploits input-side saliency signals: LLaVA-OneVision-2~\cite{an2026llavaov2} uses codec-stream tokenization to allocate visual capacity according to bit-cost dynamics and motion residuals. These approaches show that redundancy reduction is essential for practical long-video MLLMs, but they typically rely on frame-level selection, downstream token routing, task-conditioned sparsification, or codec-specific preprocessing. VideoChat3 explores a complementary route: it inflates a pretrained image ViT into an I3D-ViT that performs chunk-wise spatiotemporal self-attention and temporal pooling before the visual tokens enter the LLM. By compressing local temporal redundancy early while preserving motion-aware evidence, VideoChat3 shifts computation away from the quadratic LLM context and improves scalability for high-frame-count video inputs.

\paragraph{Data-Centric and Open Video-Language Training.}
High-quality video-language data has become a central driver of Video MLLM performance. Video-ChatGPT~\cite{maaz2023videochatgpt}, VideoChat~\cite{li2023videochat}, ShareGPT4Video~\cite{chen2024sharegpt4video}, and LLaVA-Video~\cite{llava_video} established scalable pipelines for video instruction data, often by converting captions, QA pairs, or model-generated annotations into conversational supervision. LLaVA-OneVision-2~\cite{an2026llavaov2} further scales open supervision with large re-captioned video and spatial corpora, while Molmo2~\cite{molmo2} highlights the importance of open data and introduces fully open video datasets for dense captioning, QA, pointing, and tracking without relying on closed VLM distillation. Qwen3-VL~\cite{qwen3vl} demonstrates the value of broad multimodal pretraining, long-context training, explicit timestamp alignment, and large-scale post-training, echoing a broader move toward large-scale video representation pretraining~\cite{wang2025internvideonext}. However, strong models are often either proprietary, open-weight only, or released without complete training data and construction pipelines. VideoChat3 follows the data-centric view but focuses specifically on reproducible video understanding: it combines re-annotated academic video data, synthesized long-video supervision, and online streaming QA data, and releases the model weights, code, training recipe, datasets, and data construction pipeline to make the full training stack inspectable and reusable.

\paragraph{Temporal Grounding and Streaming Video Understanding.}
Beyond holistic video QA, recent work has increasingly studied whether Video MLLMs can localize evidence in time and interact with videos as they unfold.
Early video temporal grounding methods~\cite{timechat,huang2024vtimellm,huang2024lita,guo2024vtgllm,wang2024groundedvideollm,zhu2024dualdetr} enhance fine-grained localization through timestamp-aware representations, temporal streams, and boundary-sensitive objectives; TimeLens~\cite{timelens} systematically revisits data quality, evaluation, and training recipes, while TimeLens2~\cite{zhu2026timelens2} further curates large-scale, high-quality multi-span grounding data across diverse video settings and introduces a distance-aware reward to provide more reliable signals for reinforcement learning.
VideoChat-R1~\cite{videochat-r1} and Video-R1~\cite{feng2025videor1} further improve spatiotemporal perception and video reasoning with verifiable reward signals. In parallel, online video understanding has emerged as a realistic setting in which models must process partial observations, maintain memory, and respond at the right moment. Benchmarks and systems such as OVBench~\cite{ovbench-videochatonline}, ODVBench~\cite{odvbench-streamforest}, OVOBench~\cite{ovobench}, StreamingBench~\cite{streamingbench}, VideoLLM-Online~\cite{videollmonline}, VideoChat-Online~\cite{ovbench-videochatonline}, Flash-VStream~\cite{flashvstream}, StreamBridge~\cite{streambridge}, StreamingVLM~\cite{xu2025streamingvlm}, and StreamForest~\cite{odvbench-streamforest} expose the gap between offline video QA and real-time streaming interaction. VideoChat3 connects these two directions by training on long-form temporal grounding, dense captioning, long-video QA, and proactive streaming QA. Its adaptive frame resolution further lets the model spend more visual budget on response-critical moments and less on routine intervals, providing a unified framework for offline and online video understanding.

\section{Conclusion}
\label{sec:conclusion}

In this work, we have introduced VideoChat3, a fully open, efficient, and generalist Video MLLM that addresses three core limitations of existing open-source video models: limited cross-domain generalization, high computational overhead, and incomplete openness of training assets. Benefiting from two complementary designs targeting efficiency and effectiveness, VideoChat3 strikes a strong balance between broad generalization and computational efficiency: the proposed Inflated 3D Vision Transformer and Adaptive Frame Resolution for Streaming Video Perception enable efficient spatiotemporal modeling and reduce video processing costs in both training and inference, while a scalable data synthesis pipeline curates three high-quality datasets covering general, long-form, and streaming scenarios to enhance cross-domain generalization. Experiments show that with only 4B parameters, VideoChat3 outperforms prior open-source models of equal or even larger scales across general, long-form, and streaming benchmarks with higher running efficiency. By fully releasing all model and training resources, we provide a fully reproducible research foundation for the open-source community, paving the way for broader advances in practical real-world video understanding systems.

\newpage

\bibliographystyle{unsrtnat}
\bibliography{refs}

\appendix
\newpage

\section{Appendix}

\subsection{Implementation Details}
\label{appendix:evaluation_configurations}

\paragraph{Offline video understanding.}
We evaluate VideoChat3-4B on the offline
video understanding benchmarks using the benchmark-specific configurations
summarized in \Cref{tab:eval_config_offline}. These configurations specify
the video sampling rate, maximum number of sampled frames, maximum per-frame
pixel budget, and total visual-token budget. For a consistent comparison, we
apply the same configurations to the baseline models whenever they are
supported. If a model cannot accommodate a particular configuration due to
limitations on the number of frames, spatial resolution, or visual-token
budget, we follow the default evaluation setting recommended by its official
implementation.

\begin{table*}[htbp]
    \centering
    \renewcommand{\arraystretch}{1.05}
    \setlength{\tabcolsep}{8pt}

    \caption{\textbf{Evaluation configurations for offline video understanding benchmarks.}
    ``Sampling Rate'' denotes the video sampling frequency, ``Max. Frames''
    denotes the maximum number of sampled frames, ``Pixel Budget / Frame''
    denotes the maximum number of input pixels allocated to each frame, and
    ``Visual Token Budget'' denotes the maximum total number of visual tokens.}
    \label{tab:eval_config_offline}

    \begin{tabular}{@{}l
        >{\columncolor{vc3tableLight}}c
        >{\columncolor{vc3tableLight}}c
        >{\columncolor{vc3tableMid}}c
        >{\columncolor{vc3tableMid}}c@{}}
        \toprule
        \multirow{2}{*}{\textbf{Benchmark}}
        & \multicolumn{2}{c}{\textbf{Temporal Sampling}}
        & \multicolumn{2}{c}{\textbf{Visual Budget}} \\
        \cmidrule(lr){2-3}
        \cmidrule(lr){4-5}
        & \textbf{Sampling Rate}
        & \textbf{Max. Frames}
        & \textbf{Pixel Budget / Frame}
        & \textbf{Visual Token Budget} \\
        \midrule

        \multicolumn{5}{@{}l}{\textbf{\textit{Temporal Perception}}} \\
        TOMATO~\cite{tomato}
        & 8 FPS & 2,048 & $448^2$ & 80K \\
        MotionBench~\cite{motionbench}
        & 8 FPS & 2,048 & $448^2$ & 80K \\
        TVBench~\cite{tvbench}
        & 8 FPS & 2,048 & $448^2$ & 80K \\
        TempCompass~\cite{tempcompass}
        & 8 FPS & 2,048 & $448^2$ & 80K \\
        
        \midrule
        \multicolumn{5}{@{}l}{\textbf{\textit{Long-Video Understanding}}} \\
        Video-MME~\cite{videomme}
        & 2 FPS & 1,024 & $448^2$ & 80K \\
        LVBench~\cite{lvbench}
        & 2 FPS & 2,048 & $448^2$ & 80K \\
        VideoEval-Pro~\cite{videoeval-pro}
        & 2 FPS & 2,048 & $448^2$ & 80K \\
        
        \midrule
        \multicolumn{5}{@{}l}{\textbf{\textit{Video Reasoning}}} \\
        Video-MME-v2~\cite{Video-MME-v2}
        & 1 FPS & 512 & $448^2$ & 80K \\
        Video-MMMU~\cite{Video-mmmu}
        & 2 FPS & 512 & $768^2$ & 80K \\
        MMVU~\cite{mmvu}
        & 2 FPS & 2,048 & $768^2$ & 80K \\
        Minerva~\cite{minerva}
        & 2 FPS & 2,048 & $448^2$ & 80K \\
        
        \midrule
        \multicolumn{5}{@{}l}{\textbf{\textit{Temporal Grounding}}} \\
        TimeLens~\cite{timelens}
        & 4 FPS & 2,048 & $640^2$ & 128K \\
        VUE-TR V1~\cite{vidi}
        & 1 FPS & 2,048 & $480^2$ & 128K \\
        VUE-TR V2~\cite{vidi2}
        & 1 FPS & 2,048 & $480^2$ & 128K \\
        MomentSeeker~\cite{momentseeker}
        & 2 FPS & 2,048 & $480^2$ & 128K \\

        \bottomrule
    \end{tabular}
\end{table*}

\paragraph{Online video understanding.}
The evaluation configurations for the online video benchmarks are summarized
in \Cref{tab:online_perception_memory_configurations}
and \Cref{tab:online_proactive_configurations}. We evaluate VideoChat3-4B and
Qwen3-VL-4B using the configurations reported in these tables. For the
remaining online video models, most results are taken from their original
papers. When an officially reported result is unavailable, we evaluate the
model using a sampling rate of no more than 2 FPS, taking into account its
supported video processing capacity.

\Cref{tab:online_perception_memory_configurations} reports the
configurations for the perception-and-memory benchmarks, including the video
sampling rate, maximum number of retained frames, maximum per-frame pixel
budget, and total visual-token budget. In particular, OVOBench processes the
video using a 32-frame sliding window rather than retaining all sampled frames
simultaneously. For ODVBench and OVBench, during evaluation, we modify the
required bounding-box output range from $[0,1]$ to $[0,1000]$.
\Cref{tab:online_proactive_configurations} reports the
configurations for the proactive-response benchmarks. 
For VideoChat3 specifically, the table specifies the sampling rate, maximum number of
frames, per-frame visual-token budgets under the low- and high-resolution
perception states, and the high-resolution window size. The latter is defined as
the number of consecutive frames processed at high resolution after the model
predicts the \emph{Standby} state.
For OVOBench, we report both the overall score and the task-average score,
where ``Task Avg.'' denotes the average performance over all subtasks within
each benchmark component. For ProactiveVQA, we report the Proactive Area Under
Curve (PAUC) score with the default setting \(\omega=0.5\).

\begin{table*}[htbp]
    \centering
    \renewcommand{\arraystretch}{1.08}
    \setlength{\tabcolsep}{8pt}

    \caption{\textbf{Evaluation configurations for online perception-and-memory benchmarks.}
    ``Max. Frames'' denotes the maximum number of frames processed for each video
    sample. For OVOBench, \(32^{*}\) denotes the size of the sliding window rather
    than the maximum number of frames sampled from the entire video.
    ``Pixel Budget / Frame'' denotes the maximum number of input pixels allocated
    to each frame, and ``Visual Token Budget'' denotes the maximum total number of
    visual tokens. A dash indicates that no explicit constraint is applied to the
    corresponding setting.}
    \label{tab:online_perception_memory_configurations}

    \begin{tabular}{@{}l
        >{\columncolor{vc3OnlineLight}}c
        >{\columncolor{vc3OnlineLight}}c
        >{\columncolor{vc3OnlineMid}}c
        >{\columncolor{vc3OnlineMid}}c@{}}
        \toprule
        \multirow{2}{*}{\textbf{Benchmark}}
        & \multicolumn{2}{c}{\textbf{Temporal Sampling}}
        & \multicolumn{2}{c}{\textbf{Visual Budget}} \\
        \cmidrule(lr){2-3}
        \cmidrule(lr){4-5}
        & \textbf{Sampling Rate}
        & \textbf{Max. Frames}
        & \textbf{Pixel Budget / Frame}
        & \textbf{Visual Token Budget} \\
        \midrule

        OVBench~\cite{ovbench-videochatonline}
        & 2 FPS
        & 4,096
        & $448^2$
        & 80K \\
        
        ODVBench~\cite{odvbench-streamforest}
        & 4 FPS
        & 4,096
        & $448^2$
        & 80K \\
        
        OVOBench~\cite{ovobench}
        & 2 FPS
        & 32\textsuperscript{*}
        & $448^2$
        & 80K \\
        
        StreamingBench~\cite{streamingbench}
        & 4 FPS
        & 4,096
        & $448^2$
        & 80K \\
        
        River~\cite{river}
        & 2 FPS
        & 512
        & \VCBenchNew
        & \VCBenchNew \\
        \bottomrule
    \end{tabular}
\end{table*}

\begin{table*}[htbp]
    \centering
    \renewcommand{\arraystretch}{1.08}
    \setlength{\tabcolsep}{6pt}

    \caption{\textbf{Evaluation configurations for online proactive-response benchmarks.}
    ``Low Tokens / Frame'' and ``High Tokens / Frame'' denote the maximum
    per-frame visual-token budgets under low- and high-resolution perception,
    respectively. ``High-Budget Frames after Standby'' denotes the number of
    subsequent frames processed with the high token budget after a
    \emph{Standby} prediction.}
    \label{tab:online_proactive_configurations}

    \begin{tabular}{@{}l
        >{\columncolor{vc3OnlineLight}}c
        >{\columncolor{vc3OnlineLight}}c
        >{\columncolor{vc3OnlineMid}}c
        >{\columncolor{vc3OnlineMid}}c
        >{\columncolor{vc3OnlineMid}}c@{}}
        \toprule
        \multirow{2}{*}{\textbf{Benchmark}}
        & \multicolumn{2}{c}{\textbf{Temporal Sampling}}
        & \multicolumn{3}{c}{\textbf{Adaptive Visual Allocation}} \\
        \cmidrule(lr){2-3}
        \cmidrule(lr){4-6}
        & \textbf{Sampling Rate}
        & \textbf{Max. Frames}
        & \shortstack{\textbf{Low Tokens}\\\textbf{/ Frame}}
        & \shortstack{\textbf{High Tokens}\\\textbf{/ Frame}}
        & \shortstack{\textbf{High-Budget Frames}\\
                      \textbf{after Standby}} \\
        \midrule

        OVO-Timing~\cite{ovotiming-emgarde}
        & 4 FPS
        & 128
        & 64
        & 256
        & 3 \\
        
        ProactiveVQA~\cite{proactivevideoqa}
        & 4 FPS
        & 128
        & 128
        & 512
        & 3 \\

        \bottomrule
    \end{tabular}
\end{table*}

\subsection{Demo}

\begin{figure}[htbp]
  \centering
  \includegraphics[width=\linewidth]{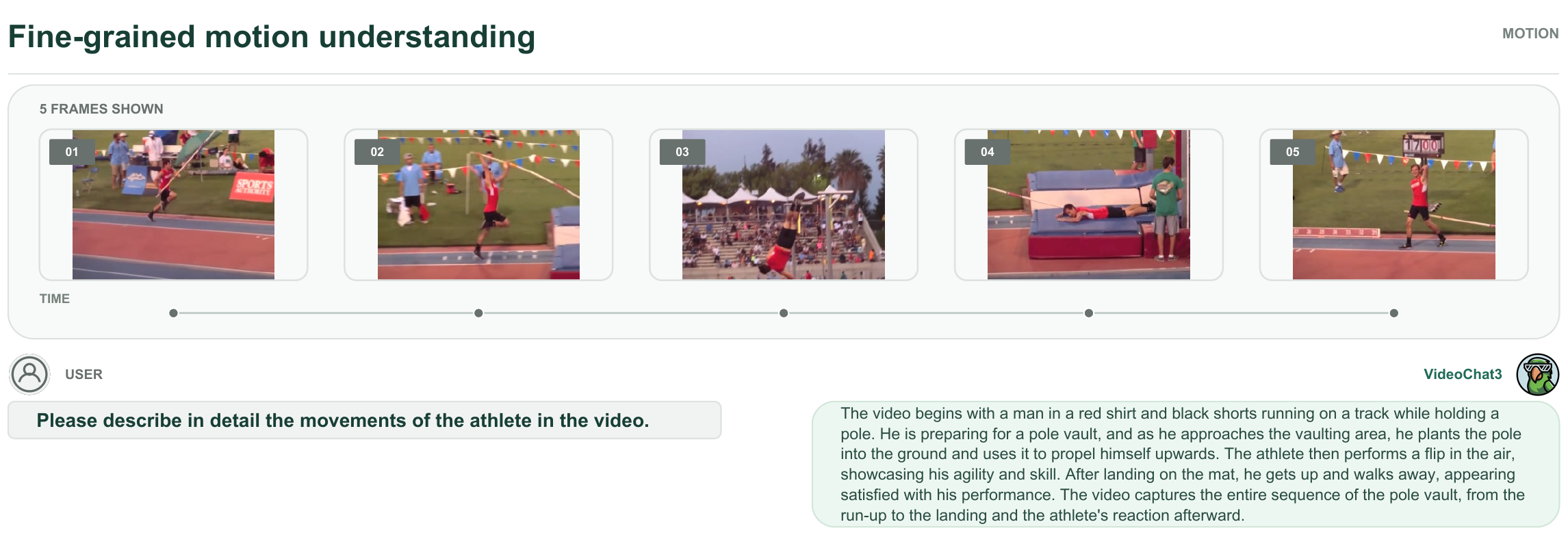}
  \caption{\textbf{Qualitative example of fine-grained video captioning.}
  VideoChat3 describes the complete pole-vault sequence, covering the run-up, pole planting, takeoff, bar clearance, and landing.}
  \label{fig:demo_motion}
\end{figure}

\begin{figure}[htbp]
  \centering
  \includegraphics[width=\linewidth]{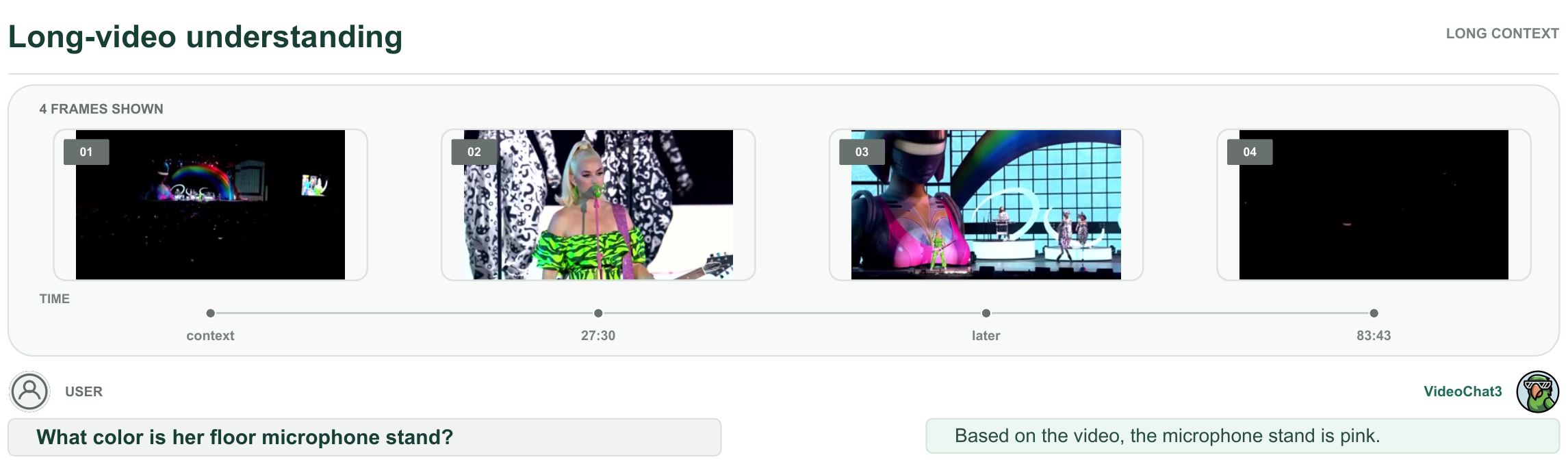}
  \caption{\textbf{Qualitative example of long-video question answering.}
  VideoChat3 retrieves a specific visual detail from an extended video and correctly identifies the color of the floor microphone stand.}
  \label{fig:demo_longvideo}
\end{figure}

\begin{figure}[htbp]
  \centering
  \includegraphics[width=\linewidth]{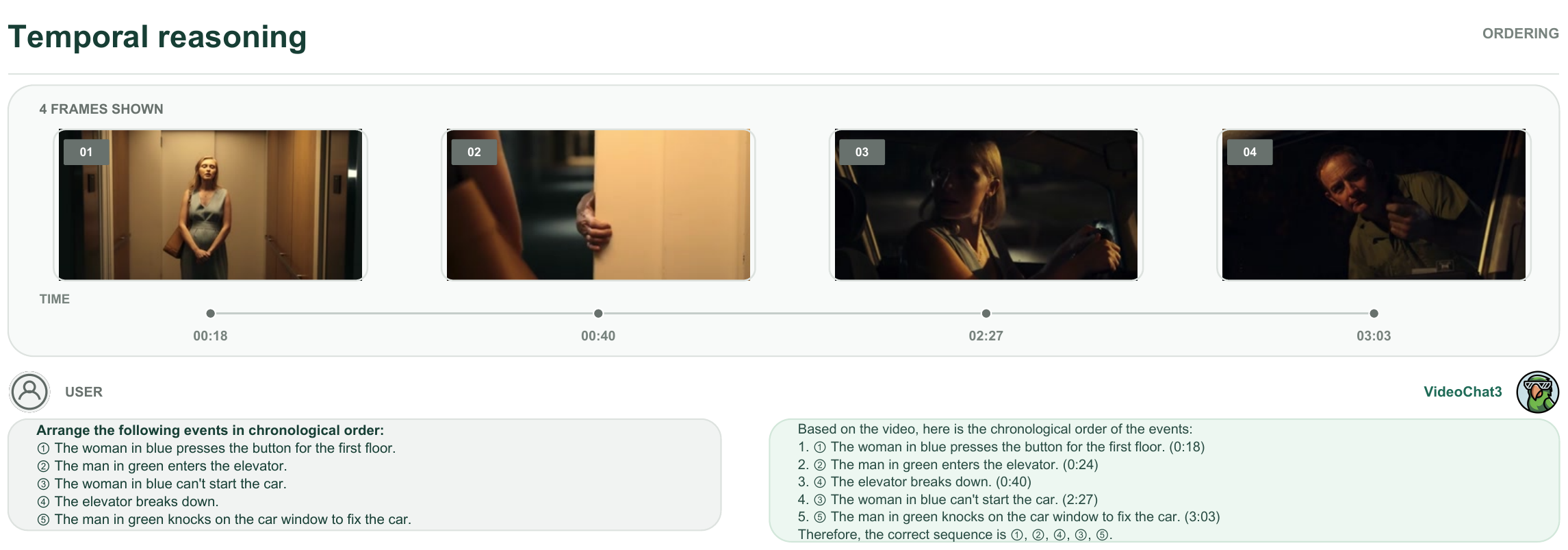}
  \caption{\textbf{Qualitative example of temporal video reasoning.}
  VideoChat3 recognizes events distributed across the video and arranges them in chronological order.}
  \label{fig:demo_reasoning}
\end{figure}

\begin{figure}[htbp]
  \centering
  \includegraphics[width=\linewidth]{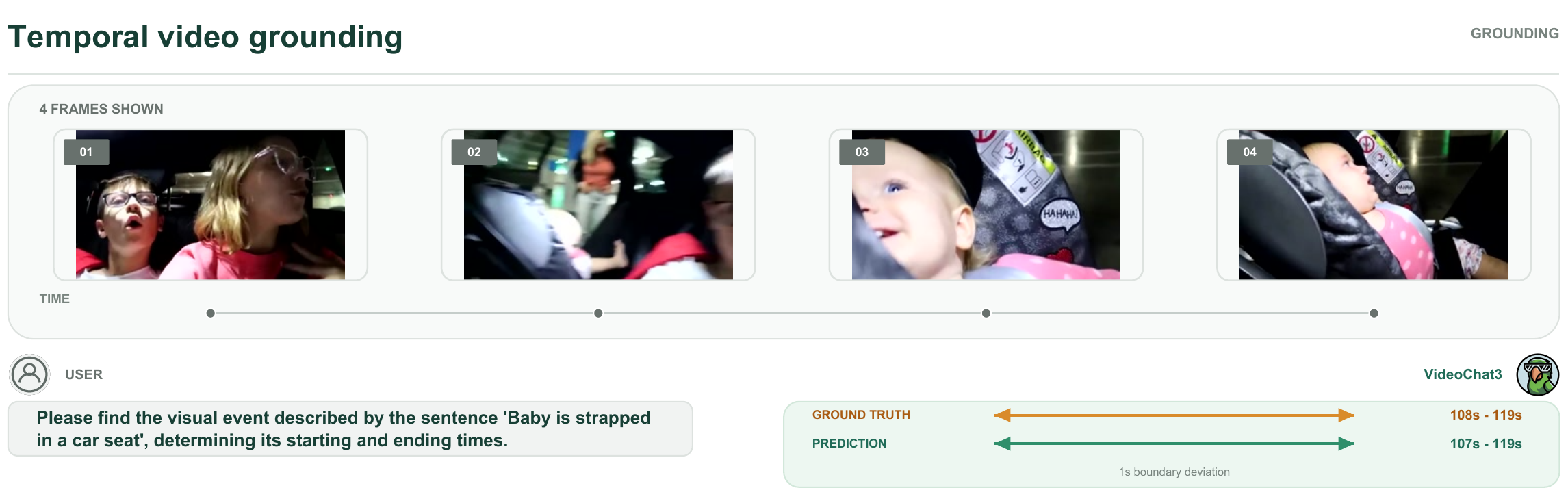}
  \caption{\textbf{Qualitative example of temporal video grounding.}
  Given a language description, VideoChat3 accurately localizes the corresponding event, with only a minor boundary deviation from the ground-truth interval.}
  \label{fig:demo_tvg}
\end{figure}

\begin{figure}[htbp]
  \centering
  \includegraphics[width=\linewidth]{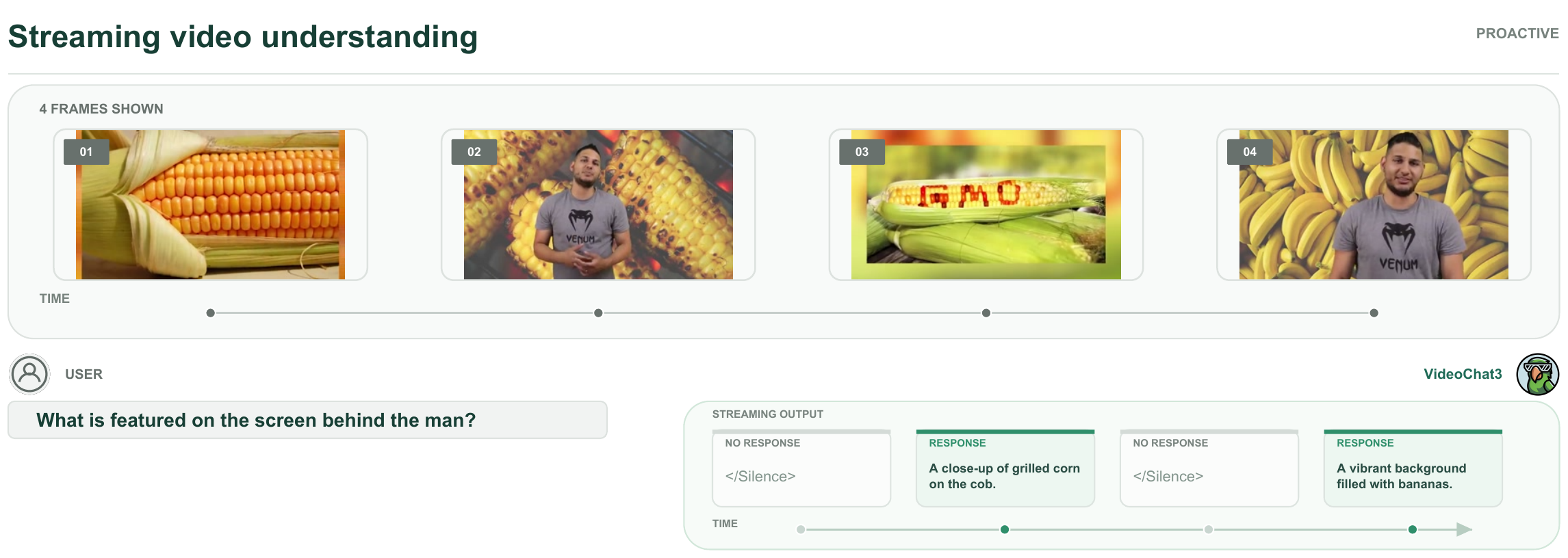}
  \caption{\textbf{Qualitative example of online proactive response.}
  During streaming perception, VideoChat3 remains silent when relevant evidence is absent and produces timely responses once informative visual content appears.}
  \label{fig:demo_proactive}
\end{figure}

\begin{figure}[htbp]
  \centering
  \includegraphics[width=\linewidth]{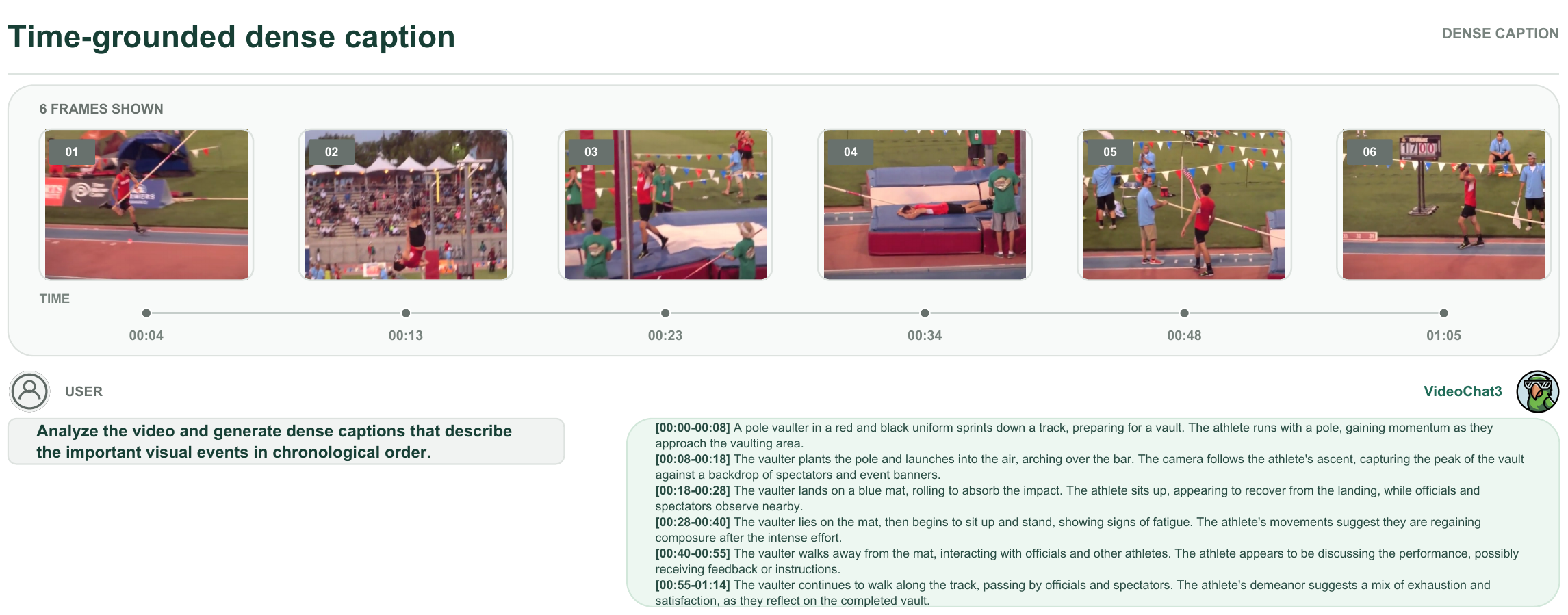}
  \caption{\textbf{Qualitative example of dense video captioning.}
  VideoChat3 identifies multiple events throughout the video and generates temporally coherent descriptions that capture its evolving content.}
  \label{fig:demo_dense_caption}
\end{figure}

We provide additional qualitative demonstrations of the broad video understanding capabilities of VideoChat3 in
\Cref{fig:demo_motion,fig:demo_dense_caption,fig:demo_longvideo,fig:demo_reasoning,fig:demo_tvg,fig:demo_proactive}.
These examples cover fine-grained motion captioning, dense video captioning, long-video question answering, temporal reasoning, temporal video grounding, and online proactive response.
In the fine-grained motion example, VideoChat3 captures subtle action transitions and describes the complete progression of a complex movement.
For dense video captioning, the model recognizes multiple events distributed throughout the video and produces temporally coherent descriptions of the evolving content.
The long-video example demonstrates its ability to retrieve a specific visual detail from an extended temporal context, while the reasoning example shows that it can identify and chronologically organize events occurring at different timestamps.
At a finer temporal granularity, VideoChat3 accurately localizes a language-described event along the video timeline.
Finally, in the streaming setting, the model selectively remains silent when relevant visual evidence is absent and proactively responds once informative content becomes available.

\end{document}